\documentclass{article}

\usepackage[final]{neurips_2024}

\usepackage[utf8]{inputenc} 
\usepackage[T1]{fontenc}    
\usepackage{hyperref}       
\usepackage{url}            
\usepackage{booktabs}       
\usepackage{amsfonts}       
\usepackage{nicefrac}       
\usepackage{microtype}      
\usepackage{xcolor}         
\usepackage{array}
\usepackage{microtype}
\usepackage{graphicx}
\usepackage{booktabs} 
\usepackage{subfig}
\usepackage{algorithm, setspace} 
\usepackage{algpseudocode} 
\usepackage{multirow}
\usepackage{arydshln}
\usepackage{comment}

\usepackage{hyperref}
\usepackage{wrapfig}

\usepackage{amsmath}
\usepackage{amssymb}
\usepackage{mathtools}
\usepackage{amsthm}

\usepackage{makecell}

\usepackage[capitalize,noabbrev]{cleveref}

\theoremstyle{plain}

\theoremstyle{definition}

\theoremstyle{remark}

\usepackage{tabularx}

\usepackage[textsize=tiny]{todonotes}

\title{DropBP: Accelerating Fine-Tuning of Large Language Models by Dropping Backward Propagation}

\author{%
  \textbf{Sunghyeon Woo}\textsuperscript{1}\footnotemark[1]\,\,\,\footnotemark[2] \qquad
  \textbf{Baesung Park}\textsuperscript{2}\footnotemark[1] \qquad
  \textbf{Byeongwook Kim}\textsuperscript{2} \qquad
  \textbf{Minjung Jo}\textsuperscript{2} \\
  \textbf{Se Jung Kwon}\textsuperscript{2} \qquad
  \textbf{Dongsuk Jeon}\textsuperscript{1} \qquad
  \textbf{Dongsoo Lee}\textsuperscript{2} \\
  Seoul National University\textsuperscript{1} \qquad NAVER Cloud\textsuperscript{2} \\
}

\begin{document}

\maketitle

\footnotetext[1]{Equal contribution}
\footnotetext[2]{Intern at NAVER Cloud}

\begin{abstract}
  Large language models (LLMs) have achieved significant success across various domains. However, training these LLMs typically involves substantial memory and computational costs during both forward and backward propagation. While parameter-efficient fine-tuning (PEFT) considerably reduces the training memory associated with parameters, it does not address the significant computational costs and activation memory. In this paper, we propose Dropping Backward Propagation (DropBP), a novel approach designed to reduce computational costs and activation memory while maintaining accuracy. DropBP randomly drops layers during backward propagation, which is essentially equivalent to training shallow submodules generated by undropped layers and residual connections. Additionally, DropBP calculates the sensitivity of each layer to assign an appropriate drop rate, thereby stabilizing the training process. DropBP is not only applicable to full fine-tuning but can also be orthogonally integrated with all types of PEFT by dropping layers during backward propagation. Specifically, DropBP can reduce training time by 44\% with comparable accuracy to the baseline, accelerate convergence to the same perplexity by 1.5$\times$, and enable training with a sequence length 6.2$\times$ larger on a single NVIDIA-A100 GPU. Furthermore, our DropBP enabled a throughput increase of 79\% on a NVIDIA A100 GPU and 117\% on an Intel Gaudi2 HPU. The code is available at {\href{https://github.com/WooSunghyeon/dropbp}{https://github.com/WooSunghyeon/dropbp}}.
\end{abstract}

\section{Introduction}\label{Section: Introduction}
Since the advent of the transformer architecture \cite{Vaswani2017-transformer}, the field of language modelling has experienced dramatic advancements. Especially, following the scaling laws \cite{kaplan-scaling-laws1, Hoffmann-scaling-laws2}, the development of Large Language Models (LLMs) \cite{Brown-gpt3, openai-gpt4, anil-gemini, Touvron-llama, Touvron-llama2, llama3-8b} continues with the aim of achieving or outperforming human capabilities. However, tremendously increasing the size of the model results in significant costs for training from scratch. An alternative approach for developing high-capability LLMs without the costly pretraining on extensive datasets is instruction tuning \cite{Wei-FLAN, taori-alpaca, Zhou-lima, databricks-dolly}. This method fine-tunes well-trained foundation models on relatively small instruction-following datasets, enabling the models to better understand and follow prompts. 

While fine-tuning Large Language Models (LLMs) on instruction-following datasets is more cost-effective than training from scratch, it still requires substantial memory for parameters and activations, along with significant floating-point operations (FLOPs). In this context, Parameter-Efficient Fine-Tuning (PEFT) techniques \cite{Hu-LoRA, Zhang-LLaMA-Adapter, Gao-LLaMA-ADAPTER-V2} effectively reduce the memory required for parameter gradients and optimizer states by freezing pretrained weights and selectively training newly added modules. Moreover, when combined with quantization methods \cite{kwon-alpatune, Dettmers-QLoRA, Xu-QALoRA, Kim-PEQA}, these techniques can further significantly decrease the memory requirements for parameters.

\begin{figure}[!t]
\centering
\subfloat[Training time per sample of LLaMA2-7B. \label{fig: Full-FT vs LoRA vs Full-FT+DropBP training time.}]{\includegraphics[width=.5\textwidth]{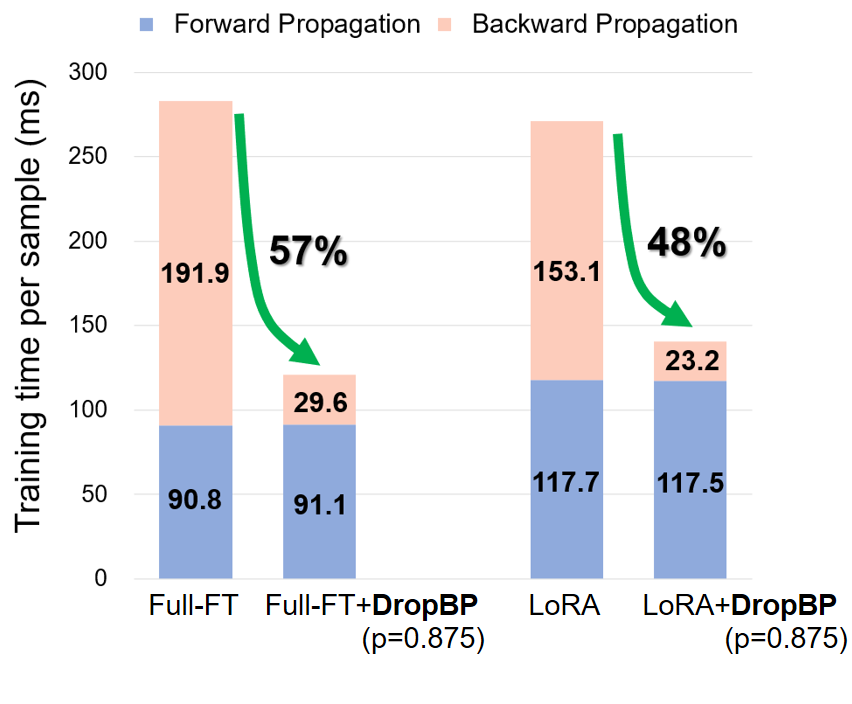}}
\subfloat[Available max sequence length of LLaMA2-70B.\label{fig: enable long sequence length.}]{\includegraphics[width=.5\textwidth]{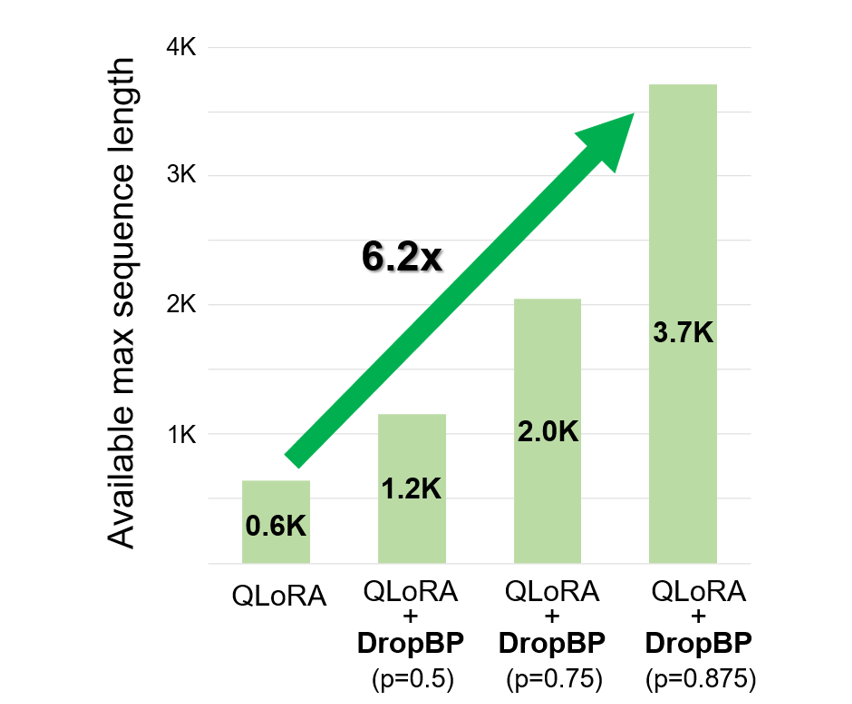}}
\caption{Performance enhancements in fine-tuning large language models using DropBP when the $p$ represents the target average drop rate for backward propagation: (a) Training time per sample for fine-tuning LLaMA2-7B with DropBP, at a sequence length of 512 and a micro batch size of 2. (b) Available max sequence length for fine-tuning LLaMA2-70B with DropBP, at a micro batch size of 1 on an NVIDIA-A100 GPU.}
\label{fig: benefit of DropBP}
\end{figure}

While Parameter-Efficient Fine-Tuning (PEFT) has successfully reduced memory associated with parameters, two significant challenges remain for efficient fine-tuning: computational cost and activation memory, both of which are linked to the backpropagation \cite{rumelhart1-backprop}. First, fine-tuning Large Language Models (LLMs) using a backpropagation requires substantial floating-point operations (FLOPs). Specifically, the backpropagation algorithm necessitates forward propagation to calculate outputs and backward propagation to compute gradients for inputs and parameters. Notably, backward propagation demands twice the computational operations compared to forward propagation, thus becoming the primary bottleneck. Second, all intermediate outputs (i.e., activations) generated during forward propagation must be stored for compute in backward propagation. This activations consume considerable memory, which becomes especially critical when training LLMs on long sequence contexts \cite{chen-longolora, xiong-long-context1}.

In this paper, we introduce Dropping Backward Propagation (DropBP), an efficient fine-tuning algorithm for LLMs that significantly reduces computational costs and activation memory. DropBP randomly drops layers during backward propagation, which is essentially equivalent to training shallow submodules generated by undropped layers and residual connections. As a result, these undropped layers no longer require FLOPs and activation memory during backward propagation. Additionally, DropBP calculates the sensitivity of each layer, an indicator of its impact on the total training process, to adjust drop rate. This careful calibration of drop rate according to layer sensitivity ensures more stable training. This DropBP algorithm can be seamlessly integrated with any PEFT algorithm, operating orthogonally by simply dropping layers during backward propagation.

We implemented DropBP as an easy-to-integrate PyTorch library \cite{pszke-pytorch}, requiring only minimal changes to the existing training codes. In experiments, DropBP successfully reduces training time as shown in Fig. \ref{fig: Full-FT vs LoRA vs Full-FT+DropBP training time.}, maintaining comparable accuracy on the MMLU \cite{Hendrycks-MMLU} and commonsense reasoning tasks \cite{Bisk-PIQA, Zellers-Hellaswag, Bhakthavatsalam-Arc-C, Mihaylov-OBQA, Sakaguch-Winogrande}. The DropBP also accelerated the convergence of the same perplexity by 1.5$\times$ in LLaMA2-70B \cite{Touvron-llama2}. Moreover, DropBP substantially decreases activation memory, increasing an available maximum sequence length by up to 6.2 $\times$ in LLaMA2-70B on a single NVIDIA A100 GPU \cite{a100-svedin}, as shown in Fig. \ref{fig: enable long sequence length.}. Finally, our DropBP increases training throughput by up to 79\% and 117\% on a single NVIDIA A100 GPU and Intel Gaudi2 HPU \cite{gaudi2-interl}, respectively, when fully fine-tuning LLaMA3-8B \cite{llama3-8b}. In summary, the main contributions of our paper are: 

\begin{itemize}
\item We propose DropBP, an efficient fine-tuning algorithm that randomly drops backward propagation based on layer sensitivity.
\item We implemented DropBP as a user-friendly PyTorch extension with a straightforward API for ease of use, making it easily applicable to existing training codes.
\item DropBP reduces training time by 44\% with comparable accuracy, increases convergence speed by 1.5$\times$ , increases the available maximum sequence length up to 6.2$\times$, and enhances training throughput up to 117\%. 
\end{itemize}

\section{Background \& Motivation}\label{Section: Background}

\paragraph{Backpropagation}
Backpropagation \cite{kelly-backprop}, a core algorithm for training deep neural networks, involves both forward and backward propagation. Specifically, the training process in the linear layer is represented as follows:

\begin{align}
 \text{Forward Prop:} \quad &\textbf{H}_{out} = \textbf{W} \times \textbf{H}_{in} \label{eq: forward propagation} \\
 \text{Backward Prop:} \quad &\nabla \textbf{H}_{in} =  \textbf{W}^{T} \times \nabla\textbf{H}_{out}  \label{eq: backward propagation} \\
  &\nabla \textbf{W} = \nabla \textbf{H}_{out} \times \textbf{H}_{in}^{T} \label{eq:parameter updates}
 \end{align}

where $\textbf{H}$ and $\textbf{W}$ represent the activations and parameters, respectively, with '$\times$' indicating matrix multiplication operation. The gradients of $\textbf{H}$ and $\textbf{W}$ are denoted by $\nabla \textbf{H}$ and $\nabla \textbf{W}$. The computational costs during forward propagation primarily arises from matrix multiplication for computing output activations by Eq. \ref{eq: forward propagation}. In backward propagation, the computational burden is primarily due to matrix multiplication for calculating input gradients by Eq. \ref{eq: backward propagation} and parameter gradients by Eq. \ref{eq:parameter updates}. Note that the computational costs of these equations are almost equal. Consequently, the FLOPs required for backward propagation including Eqs. \ref{eq: backward propagation} and \ref{eq:parameter updates} are approximately 2$\times$ as large as the FLOPs needed for forward propagation by Eq. \ref{eq: forward propagation}. Furthermore, the activations of all layers ($\textbf{H}_{in}^{T}$) must be stored in memory for use in backward propagation computations in Eq. \ref{eq:parameter updates}. Therefore, focusing on reducing the computations during backward propagation is crucial for decreasing both the overall computational costs and the activation memory.

\paragraph{Interpretation the model with residual connections}

\begin{figure}[!ht]
\centering
\subfloat[3-layers model with residual connections. \label{fig: total module.}]{\includegraphics[width=.4\textwidth]{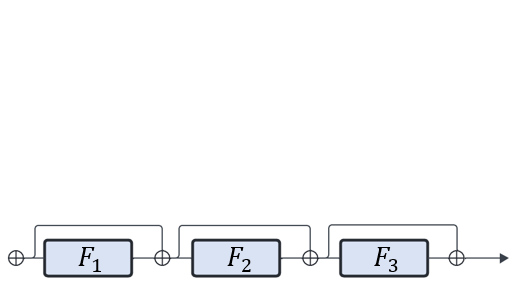}} \qquad
\subfloat[Combination of multiple submodules. \label{fig: submodule.}]{\includegraphics[width=.4\textwidth]{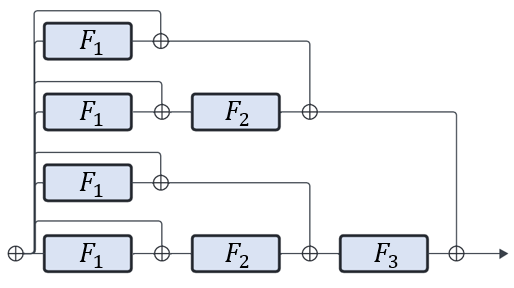}}
\caption{Interpreting the model with residual connections as a combination of multiple submodules.}
\label{fig: explanation of module}
\end{figure}

Residual connections are one of the widely used methods to address the issue of vanishing gradients \cite{He-Resnet}. Transformer \cite{Vaswani2017-transformer} also incorporates residual connections that bypass multi-head attention and feedforward networks. Networks utilizing these residual connections can be interpreted as ensembles of several submodules \cite{Veit-shallownet}. For example, if we expand the model with three residual connections as shown in Fig. \ref{fig: total module.}, it can be represented as a combination of eight submodules, as depicted in Fig. \ref{fig: submodule.} From this perspective, a network with $n$ layers can be interpreted as an ensemble of $2^{n}$ submodules \cite{Veit-shallownet}.

\section{Methodology}\label{Section: Methodology}

\subsection{Dropping Backward propagation}\label{Subsection: Dropping Backpropagation}

\begin{figure}[!ht]
\centering
\subfloat[The concept of DropBP. \label{fig: concept of dropbp}]{\includegraphics[width=.45\textwidth]{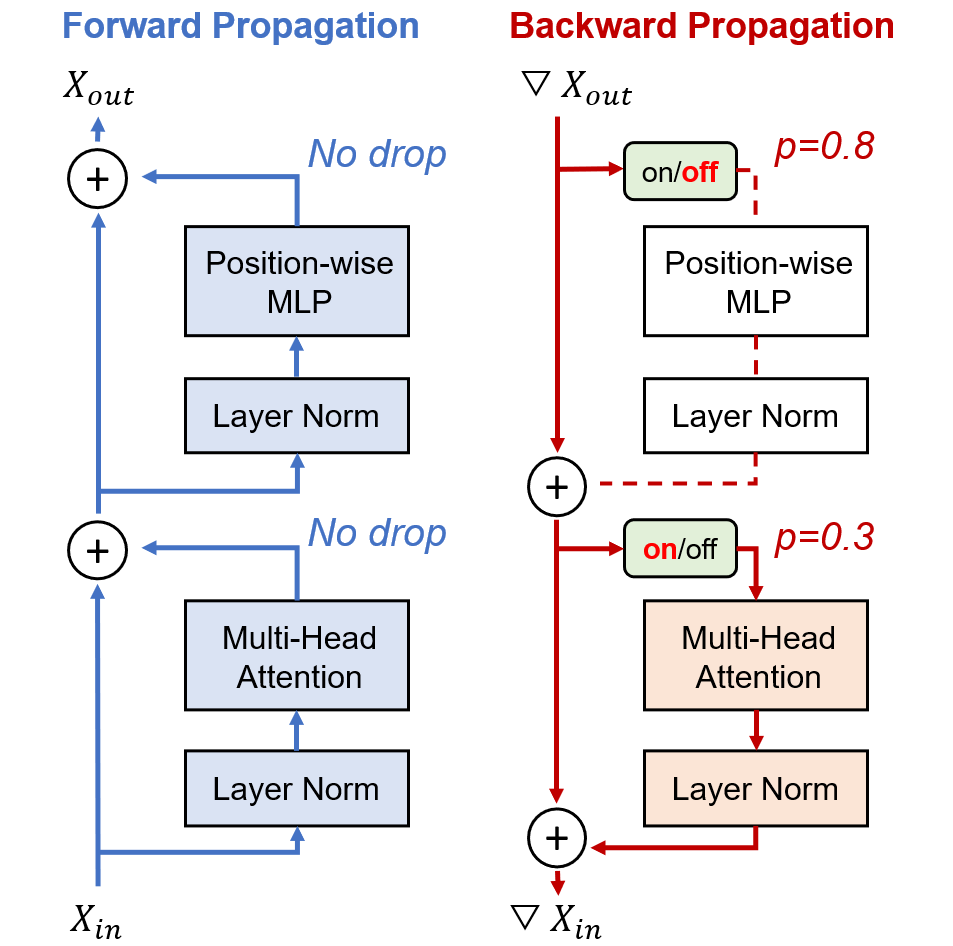}} \qquad
\subfloat[DropBP as combiation of shallow submodules.\label{fig: submodule_dropbp.}]{\includegraphics[width=.45\textwidth]{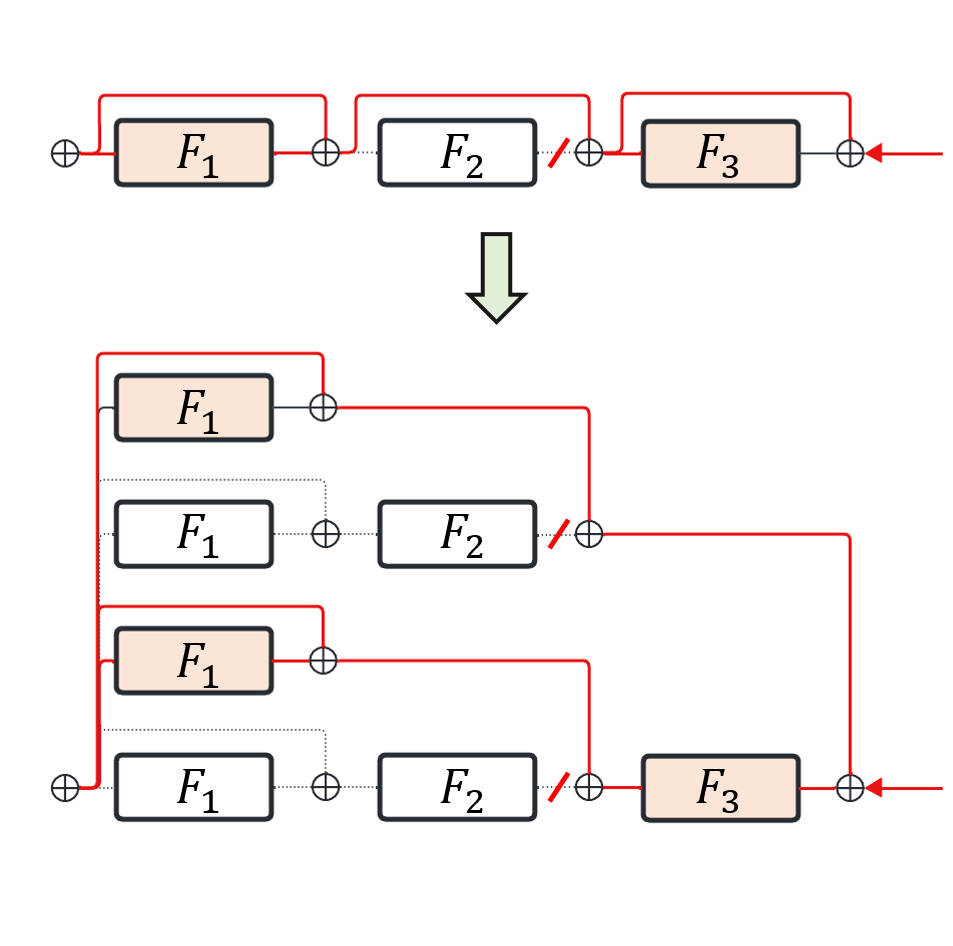}}
\caption{The overveiw of DropBP.}
\label{fig: overview of dropbp}
\end{figure}

In Section \ref{Section: Background}, we observed that the backpropagation algorithm consumes a significant amount of FLOPs and activation memory, particularly for the backward propagation. To reduce this overhead, we propose a straightforward approach: Dropping Backward Propagation (DropBP). DropBP is designed to drop layers exclusively during backward propagation, effectively reducing both FLOPs and activation memory for the dropped layers, as demonstrated in following equations:

\begin{align}
\textbf{X}_{imm} &= \textbf{X}_{in} + \mathcal{D}_{p_{i}} (f_{ATTN}(f_{LN}(\textbf{X}_{in})))  \label{eq: dropbp_attn} \\
 \textbf{X}_{out} &= \textbf{X}_{imm} + \mathcal{D}_{p_{i+1}}(f_{FFN}(f_{LN}(\textbf{X}_{imm}))) \label{eq: dropbp_ffn}
 \end{align}

Here, \(\textbf{X}_{in}\), \(\textbf{X}_{out}\), and \(\textbf{X}_{imm}\) represent the input, output, and immediate activation between the attention layer and feedforward network in a transformer block, respectively. \(f_{ATTN}\), \(f_{FFN}\), and \(f_{LN}\) denote the attention layer, feedforward network, and layer normalization of the transformer block. The DropBP layer, defined as \(\mathcal{D}_{p}(\textbf{X})\), skips backward propagation in the input $\textbf{X}$ with a probability of \(p\), while not dropping forward propagation. Following to Eqs. \ref{eq: dropbp_attn} and \ref{eq: dropbp_ffn}, backward propagation in the attention layer and feedforward network is dropped with probabilities \(p_{i}\) and \(p_{i+1}\), respectively, as shown in Fig. \ref{fig: concept of dropbp}.  

We can view the transformer as a collection of a lot of submodules composed of various modules (i.e., $f_{ATTN} \circ f_{LN}$ and $f_{FFN} \circ f_{LN}$) with residual connections, as described in Section 2. When the transformer block contains $n$ units, each including multi-head attention and a feedforward network, the model can be interpreted as an ensemble of $2^{2n}$ submodules. From this perspective, DropBP can be interpreted as training only certain shallow submodules. For example, as shown in Fig. \ref{fig: submodule_dropbp.}, if the $F_2$ layer is dropped, only the shallow submodule composed of the remaining layers is trained during backward propagation. Therefore, if the overall drop rate is \(p\), DropBP can be interpreted as training shallow submodules with the depth of \(2n(1-p)\) or less, since \(2np\) layers are dropped. We anticipate that training these smaller modules alone can effectively fine-tune well-pretrained LLMs, based on the analysis that shallow submodules have a significant impact on the overall training process as detailed in Appendix \ref{appendix: the importance of short paths in residual networks}.

\subsection{Sensitivity-based Drop Rate Allocation}\label{Subsection: Sensitivity-based Dropping Backpropagation}

In Section \ref{Subsection: Dropping Backpropagation}, we introduce DropBP, which selectively drops layers during backward propagation and trains only certain shallow submodules. In addition, we hypothesized that the significance of individual layers and the submodules encompassing these layers varies in their impact on the overall training process. Therefore, we assign different drop rate to each layer based on sensitivity, which is defined by defined by how much each layer and its encompassing submodules affect the overall training process in terms of parameter gradient. Specifically, we calculate the sensitivity of a layer by the variance in L2-norm of parameter gradients between when the backward propagation of that layer is skipped or not, inspired by GradNormVar \cite{Woo-ALAM}, a memory efficient gradient variance approximation, as below:

\begin{equation}
\label{eq: sensitivity calculation}
{S}_{l} = \sum_{i}(||\nabla\textbf{W}_{i}||_{2}-||\nabla\textbf{W}_{i}^{(l)}||_{2})^{2}
\end{equation}

where $S_{l}$ denotes of the $l$-th layer. Here, $\nabla\textbf{W}_{i}$ represnets the parameter gradient of the $i$-th layer when no layers are dropped, while $\nabla \textbf{W}^{(l)}_{i}$ denotes the parameter gradient of the $i$-th layer when the $l$-th layer is dropped during backward propagation. After calculating the sensitivity for each layer, we aim to minimize the expected sensitivities across all layers-essentially the expected gradient variance caused by DropBP-under a given FLOPs as follow: 

\begin{align}
\label{eq: allocates drop rates}
\text{min} \sum_{i} (1-p_{i}) S_{i}, \,\,\, \text{s.t.} \sum_{i}(1-p_{i})F_{i}\leq F_{t}
\end{align}

where $p_{i}$ denotes the drop rate, and $F_{i}$ denotes the FLOPs of the $i$-th layer during backward propagation. $F_{t}$ represents the target FLOPs, derived from the target average drop rate $p_{avg}$ (i.e. $F_{t} = (1-p_{avg}) \sum_{i}F_{i}$). In other words, we determine the drop rates across all layers that minimize the expected sensitivities of the model, while satisfying a given FLOPs budget, by solving Problem \ref{eq: allocates drop rates}. In practice, DropBP addresses the Prob. \ref{eq: allocates drop rates} using a simple greedy algorithm, as detailed in Section \ref{Subsection: Implementation}.

\section{Evaluation}\label{Section: Evaluation}

\subsection{Implementation and Settings}\label{Subsection: Implementation}
\begin{figure}[!ht]
\centering
\subfloat[Code for preparing a model with DropBP. \label{fig: model code}]{\includegraphics[width=.5\textwidth]{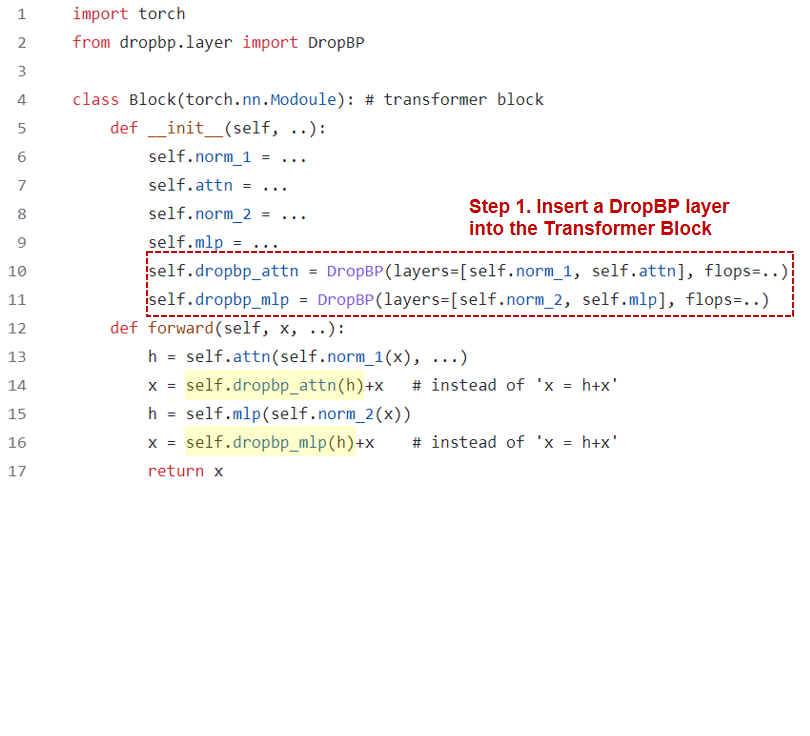}}
\subfloat[Code for training with DropBP.\label{fig: training code}]{\includegraphics[width=.5\textwidth]{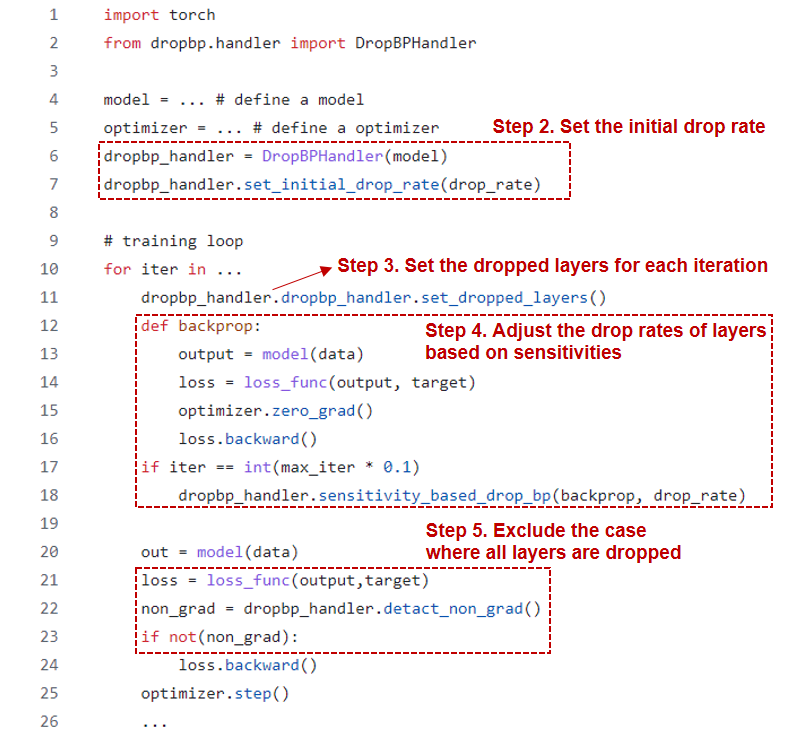}}
\caption{Code implementation for integrating DropBP.}
\label{fig: code implementation}
\end{figure}

DropBP aims to decrease the training costs in fine-tuning based on backpropagation, consequently enabling the acceleration of both full fine-tuning and parameter-efficient fine-tuning using backpropagation. To facilitate practical implementation, we developed a user-friendly DropBP library in PyTorch \cite{pszke-pytorch}, as demonstrated in Fig. \ref{fig: code implementation}. In detail, we implemented a \texttt{DropBP} layer that internally drops backward propagation in the input direction according to a given drop rate as shown in Fig. \ref{fig: model code}. The \texttt{DropBP} layer designed to initially receive the FLOPs of the layers that would be skipped ($F_{i}$) as initial value to solve Prob. \ref{eq: allocates drop rates}.

Additionally, we developed a \texttt{DropBPHandler} that automatically solve Prob. \ref{eq: allocates drop rates} by assigning varying drop rate to each layer using a simple greedy algorithm, as demonstrated in Fig. \ref{fig: training code}. Specifically, the process begins by setting the drop rate ($p_{i}$) of all layers to 0 and then gradually increases them to align with the target average drop rate ($p_{avg}$) set by the user. In each step, the drop rate for each layer is incremented by 0.1, ensuring that the increase in total expected sensitivity is kept to a minimum. We train with uniform drop rate for the initial 10\% of total iterations and then adjust the drop rate for each layer based on sensitivity when training is stable. Since sensitivity is calculated only once at the 10\% of the entire iteration, the overhead from this calculation is negligible.

We integrated our DropBP code into \textit{LitGPT} \cite{Lit-GPT} and \textit{HuggingFace} \cite{huggingface}, repositories for training and evaluating LLMs. We first fine-tuned LLaMA2-7B, 13B, and 70B \cite{Touvron-llama2} on Alpaca \cite{taori-alpaca} and Dolly \cite{databricks-dolly} datasets. The fine-tuned models were evaluated on 5-shot Massive Multitask Language Understanding (MMLU) tasks \cite{Hendrycks-MMLU} and 0-shot commonsense reasoning tasks \cite{Bisk-PIQA, Zellers-Hellaswag, Bhakthavatsalam-Arc-C, Mihaylov-OBQA, Sakaguch-Winogrande} using \textit{lm-evaluation-harness} \cite{eval-harness} library. We also fine-tunes LLaMA3-8B \cite{llama3-8b} on the Oasst1 dataset \cite{oasst1-kopf} and evaluating the model on MT-Bench task \cite{MT-Bench} using GPT4o-mini \cite{openai-gpt4} as a judge. These experiments were conducted on a single NVIDIA A100 GPU, and we measured the training time, memory usage, and convergence speed. More details about setup can be found in Appendix \ref{appendix: experimental details}.

\setcounter{table}{0}
\begin{table*}[!ht]
\centering
\renewcommand{\arraystretch}{1.1}
\caption{Test accuracy on MMLU and commonsense reasoning tasks. In DropBP, drop rate is the target average drop rate across all layers in backward propagation. Note that Full-FT stands for full fine-tuning.}
\label{tb: main results}
\resizebox{\textwidth}{!}{%
\begin{tabular}{lcc|ccccc|ccccccc}
\Xhline{3\arrayrulewidth}
 \multirow{2}{*}{\textbf{Method}} & \multirow{2}{*}{\textbf{Drop Rate}} & \multirow{2}{*}{\textbf{Dataset}} & \multicolumn{5}{c|}{\textbf{MMLU (5-shot)}}                                                        & \multicolumn{7}{c}{\textbf{Commonsense Reasoning (0-shot)}}                                                                  \\
&                                &   & \textbf{Human.} & \textbf{STEM} & \textbf{Social.} & \textbf{Other} & \textbf{Avg.} & \textbf{PIQA} & \textbf{HS} & \textbf{Arc-C} & \textbf{Arc-E} & \textbf{OBQA} & \textbf{WG} & \textbf{Avg.} \\ \Xhline{3\arrayrulewidth}
 LLaMA2-7B           & -  & -                            & 43.5   & 37.0  & 51.6  & 52.2  & 45.7                        & 79.1  & 76.0  & 46.2 & 74.5 & 44.0 & 69.3 & 64.9 \\ \Xhline{1\arrayrulewidth}
 LoRA                & -  & Alpaca                       & 42.7   & 36.3  & 50.0  & 51.2  & 44.7                        & 80.0  & 75.9  & 48.5 & 75.0 & 46.2 & 70.5 & 66.0               \\ 
 \textbf{LoRA+DropBP} & \textbf{0.5}  & Alpaca                                & 42.8   & 35.7  & 50.3  & 51.0  & \textbf{44.7}    & 79.6  & 76.0  & 48.7 & 75.5 & 46.2 & 69.6 & 65.9             \\
 LoRA+DropBP & 0.75          & Alpaca                               & 41.4   & 36.3  & 48.4  & 50.6  & 43.8             & 79.5  & 76.9  & 48.0 & 75.6 & 47.4 & 69.0 & 66.1             \\
 \textbf{LoRA+DropBP} & \textbf{0.875}  & Alpaca                              & 41.5   & 34.5  & 49.6  & 50.4  & 43.7             & 79.6  & 77.2  & 48.2 & 76.3 & 47.8 & 69.1 & \textbf{66.4}    \\ \cdashline{1-15}[1pt/1pt] 
 Full-FT & -                  & Alpaca                                  & 42.7   & 35.6  & 50.4  & 51.1  & 44.7             & 79.2  & 76.1  & 48.0 & 75.8 & 45.2 & 69.8 & 65.7               \\ 
 Full-FT+DropBP & 0.5           & Alpaca                                & 42.6   & 35.4  & 49.8  & 51.0  & 44.4             & 79.5  & 76.2  & 47.8 & 75.4 & 45.4 & 68.5 & 65.5             \\
 \textbf{Full-FT+DropBP} & \textbf{0.75}  & Alpaca                                & 42.6   & 36.7  & 51.2  & 50.9  & \textbf{45.0}    & 78.8  & 77.0  & 48.6 & 75.8 & 45.6 & 69.8 & \textbf{65.9}     \\
 Full-FT+DropBP & 0.875           & Alpaca                                & 42.7   & 35.3  & 50.7  & 51.2  & 44.7             & 79.2  & 76.9  & 46.8 & 75.3 & 46.2 & 69.3 & 65.6             \\ \cdashline{1-15}[1pt/1pt]
 \textbf{LoRA} & -                 & Dolly                       & 43.9   & 38.4  & 53.0  & 53.3  & \textbf{46.7}                        & 79.0  & 76.2  & 47.7 & 77.0 & 45.0 & 69.7 & \textbf{65.8}               \\ 
 LoRA+DropBP & 0.5  & Dolly                                & 44.0   & 36.8  & 53.1  & 53.2  & 46.4    & 79.3  & 76.3    & 47.3 & 76.5 & 44.8 & 68.8 & 65.5             \\
 LoRA+DropBP & 0.75           & Dolly                               & 43.9   & 37.0  & 52.4  & 53.1  & 46.3             & 79.4  & 76.2  & 46.2 & 75.4 & 44.8 & 68.8 & 65.1             \\
 LoRA+DropBP & 0.875           & Dolly                              & 43.6   & 36.7  & 52.3  & 53.0  & 46.1             & 79.1  & 76.1  & 45.8 & 75.3 & 44.6 & 68.4 & 64.9    \\ \cdashline{1-15}[1pt/1pt] 
 \textbf{Full-FT}  & -                 & Dolly                                 & 43.3   & 38.1  & 53.6  & 53.2  & \textbf{46.6}             & 79.3  & 76.2  & 46.8 & 76.2 & 44.2 & 68.9 & \textbf{65.3}               \\ 
 Full-FT+DropBP & 0.5   & Dolly                                & 43.4   & 37.1  & 52.9  & 53.0  & 46.2    & 79.2  & 76.2  & 46.4 & 75.6 & 44.4 & 68.8 & 65.1             \\
 Full-FT+DropBP & 0.75          & Dolly                                & 43.1   & 36.7  & 51.8  & 52.6  & 45.7             & 79.2  & 76.4  & 45.8 & 75.4 & 44.6 & 69.1 & 65.1     \\
 Full-FT+DropBP & 0.875          & Dolly                                & 42.5   & 36.8  & 52.4  & 52.4  & 45.6             & 79.2  & 76.3  & 46.2 & 75.0 & 44.8 & 69.0 & 65.1             \\

\Xhline{3\arrayrulewidth}
 LLaMA2-13B  & -          & -                                       & 52.2   & 44.1  & 62.9  & 61.5  & 54.8             & 80.6  & 79.4  & 49.5 & 77.4 & 45.6 & 72.5 & 67.5               \\ \Xhline{1\arrayrulewidth}
 LoRA        & -          & Alpaca                                  & 51.7   & 43.8  & 63.3  & 61.7  & 54.7             & 80.6  & 79.5  & 51.6 & 78.5 & 45.8 & 72.1 & 68.0               \\ 
 LoRA+DropBP & 0.5           & Alpaca                                & 52.4   & 44.2  & 63.1  & 62.0  & 55.0             & 80.7  & 79.6  & 50.9 & 78.4 & 44.8 & 71.7 & 67.7             \\
 \textbf{LoRA+DropBP} & \textbf{0.75}  & Alpaca                               & 52.1   & 44.2  & 64.1  & 61.6  & \textbf{55.1}    & 81.0  & 79.7  & 51.5 & 79.1 & 45.6 & 71.7 & \textbf{68.1}    \\
 LoRA+DropBP & 0.875          & Alpaca                              & 51.1   & 44.2  & 63.3  & 61.6  & 54.6             & 80.8  & 79.8  & 51.0 & 78.2 & 45.0 & 71.4 & 67.7             \\ \cdashline{1-15}[1pt/1pt]
 LoRA        & -          & Dolly                                  & 51.9   & 43.6  & 63.7  & 62.0  & 54.8             & 80.4  & 79.9  & 51.1 & 78.4 & 45.6 & 71.7 & 67.9               \\ 
 \textbf{LoRA+DropBP} & \textbf{0.5}  & Dolly                                & 52.4   & 44.1  & 63.4  & 62.1  & 55.1             & 80.7  & 79.9  & 50.9 & 78.5 & 45.6 & 72.3 & \textbf{68.0}             \\
 LoRA+DropBP & 0.75           & Dolly                               & 52.1   & 44.3  & 63.3  & 61.7  & 54.9             & 80.6  & 79.8  & 51.4 & 77.8 & 45.4 & 72.1 & 67.8    \\
 \textbf{LoRA+DropBP} & \textbf{0.875}  & Dolly                              & 52.8   & 43.9  & 63.4  & 61.9  & \textbf{55.1}    & 80.5  & 79.7  & 51.3 & 77.9 & 45.2 & 72.0 & 67.8             \\
\Xhline{3\arrayrulewidth}
 LLaMA2-70B   & -         & -                                       & 64.7   & 57.0    & 79.6  & 74.0  & 68.3             & 82.4   & 83.0  & 57.3 & 80.6 & 48.6 & 77.4 & 71.6             \\ \Xhline{1\arrayrulewidth} 
 QLoRA      & -           & Alpaca                                  & 64.9   & 57.0    & 79.6  & 74.0  & 68.3             & 82.8  & 83.3  & 59.6 & 82.2 & 48.4 & 78.5 & 72.5             \\ 
 QLoRA+DropBP & 0.5          & Alpaca                                & 65.8   & 56.2  & 78.8  & 73.0  & 68.1             & 83.2  & 82.9  & 60.2 & 82.2 & 48.0 & 77.9 & 72.4             \\
 QLoRA+DropBP & 0.75          & Alpaca                               & 65.0   & 55.2  & 78.8  & 73.5  & 67.7             & 83.5  & 83.1  & 58.9 & 81.3 & 48.2 & 77.3 & 72.0             \\
\textbf{QLoRA+DropBP} & \textbf{0.875} & Alpaca                              & 66.4   & 56.3  & 79.9  & 74.1  & \textbf{68.8}    & 83.4  & 83.7  & 60.1 & 81.6 & 48.6 & 78.0 & \textbf{72.6}    \\                             \cdashline{1-15}[1pt/1pt]
 QLoRA    & -             & Dolly                                  & 65.5   & 58.0    & 79.7  & 74.5  & 68.9            & 82.8  & 83.3  & 58.3 & 81.2 & 48.0 & 77.4 & 71.8             \\ 
 QLoRA+DropBP & 0.5          & Dolly                                & 65.1   & 57.2  & 79.3  & 74.1  & 68.4             & 82.8  & 83.4  & 57.8 & 81.7 & 47.6 & 78.1 & 71.9             \\
 \textbf{QLoRA+DropBP} & \textbf{0.75} & Dolly                               & 65.1   & 57.4  & 79.7  & 74.5  & \textbf{68.7}    & 82.4  & 83.5  & 58.4 & 82.0 & 48.2 & 77.8 & \textbf{72.0}     \\
QLoRA+DropBP & 0.875           & Dolly                              & 65.4   & 56.6  & 79.6  & 74.1  & 68.5             & 83.1  & 83.1  & 57.6 & 81.6 & 48.0 & 78.5 & 72.0 \\
\Xhline{3\arrayrulewidth}
\end{tabular}}
\end{table*}

\begin{table*}[!ht]
\renewcommand{\arraystretch}{1.1}
\centering
\newcolumntype{C}{>{\centering\arraybackslash}X}
\scriptsize
\caption{Training time, memory usage, and test score on MT-Bench task when fine-tuning LLaMA3-8B with DropBP on Oasst1 datasets.}
\label{tb: mt-bench}
\resizebox{\textwidth}{!}{%
\begin{tabular}{c|cc|cccccccc|c}
\hline
\textbf{Method}  & \textbf{Mem} & \textbf{Time} & \textbf{Human.} & \textbf{STEM} & \textbf{Role.} & \textbf{Extract.} & \textbf{Writing} & \textbf{Reason.} & \textbf{Coding} & \textbf{Math} & \textbf{Avg.} \\ \hline
No-tunes                           & -               & -             & 6.25                & 5.70          & 5.45              & 4.85               & 5.20             & 4.40              & 3.20           & 1.95          & 4.62          \\ \hline
LoRA                              & 57G             & 27m           & 7.00                & 6.40          & 5.70              & 5.80               & 5.30             & 4.55              & 3.25           & 2.95          & \textbf{5.12}          \\ \hline
+DropBP (p=0.5)                & 42G             & 21m           & 6.55                & 6.25          & 6.05              & 5.50               & 5.05             & 4.45              & 3.75           & 3.25          & 5.11          \\ 
+DropBP (p=0.75)                                & 36G             & 17m           & 6.75                & 5.90          & 5.80              & 5.70               & 5.35             & 4.30              & 3.60           & 3.30          & 5.09          \\ 
+DropBP (p=0.875)     & \textbf{32G}    & \textbf{16m}  & 6.60                & 6.55          & 5.90              & 5.70               & 5.70             & 3.95              & 3.40           & 2.80          & 5.08          \\ \hline
\end{tabular}
}
\end{table*}

\subsection{Main Results: Accuracy and Efficiency}\label{Subsection: main results}

\paragraph{Accuracy on MMLU and Commonsense Reasoning}

We employ DropBP to accelerate baseline fine-tuning processes, including full fine-tuning (Full-FT), LoRA \cite{Hu-LoRA}, and QLoRA \cite{Dettmers-QLoRA}, on the Alpaca and Dolly datasets. As demonstrated in Table \ref{tb: main results}, DropBP achieves accuracy comparable to the baseline, with deviations less than 1\% in all scenarios, and it even outperforms the baseline in several instances. Specifically, when DropBP is applied to fine-tune LLaMA2-7B, there is a 1\% or less decrease in 5-shot MMLU accuracy compared to the baseline, while maintaining comparable 0-shot commonsense reasoning accuracy. In contrast, for LLaMA2-13B and LLaMA2-70B, fine-tuning with DropBP results in almost no decrease in accuracy on the MMLU and commonsense reasoning tasks, even at the high drop rate of 0.875.

\paragraph{Accuracy on MT-Bench} Similar trends are observed in the MT-Bench tasks, with negligible decreases in accuracy as shown in Table \ref{tb: mt-bench}. Specifically, when fine-tuning LLaMA3-8B on the Oasst1 dataset, DropBP generates responses of comparable quality to the baseline across various generation tasks. Although scores slightly decrease as the DropBP rate increases, the model fine-tuned with a high DropBP rate of 0.875 still achieves significantly higher scores compared to non-tuned models.

\paragraph{Training Speed and Memory Usage}

\begin{table*}[!ht]
\renewcommand{\arraystretch}{1.1}
\centering
\newcolumntype{C}{>{\centering\arraybackslash}X}
\scriptsize
\caption{Time required for fine-tuning LLaMA2 models with DropBP on the Alpaca datasets when $p$ denotes the target average drop rate. The number of fine-tuning samples is 50K.}
\label{tb: training time}
\begin{tabular}{c|c|c|cccc}
\Xhline{3\arrayrulewidth}
\multirow{2}{*}{\textbf{Model}}     & \multirow{2}{*}{\textbf{Precision}} & \multirow{2}{*}{\textbf{PEFT}} & \multicolumn{4}{c}{\textbf{DropBP}}                \\ \cdashline{4-7}[1pt/1pt]
                                    &                                     &                                & \textbf{p=0 (Baseline)} & \textbf{p=0.5} & \textbf{p=0.75} & \textbf{p=0.875} \\ \Xhline{1\arrayrulewidth}
\multirow{2}{*}{\textbf{LLaMA2-7B}} & BF16-mixed                          & LoRA                           & 2.2h            & 1.7h    & 1.4h    & \textbf{1.3h}     \\ \cline{2-7}
                                    & BF16                                & Full-FT                            & 2.0h              & 1.3h  & 1.0h      & \textbf{0.8h}     \\ \Xhline{1\arrayrulewidth}
\textbf{LLaMA2-13B}                 & BF16                                & LoRA                           & 2.9h            & 2.1h   & 1.7h     & \textbf{1.5h}     \\ \Xhline{1\arrayrulewidth}
\textbf{LLaMA2-70B}                 & BF16                                & QLoRA                          & 29.6h           & 22.2h  & 18.4h   & \textbf{16.5h }    \\ \Xhline{3\arrayrulewidth}
\end{tabular}
\end{table*}

\begin{wrapfigure}{!r}{0.45\textwidth}

  \centering
  \includegraphics[width=1.0\linewidth]{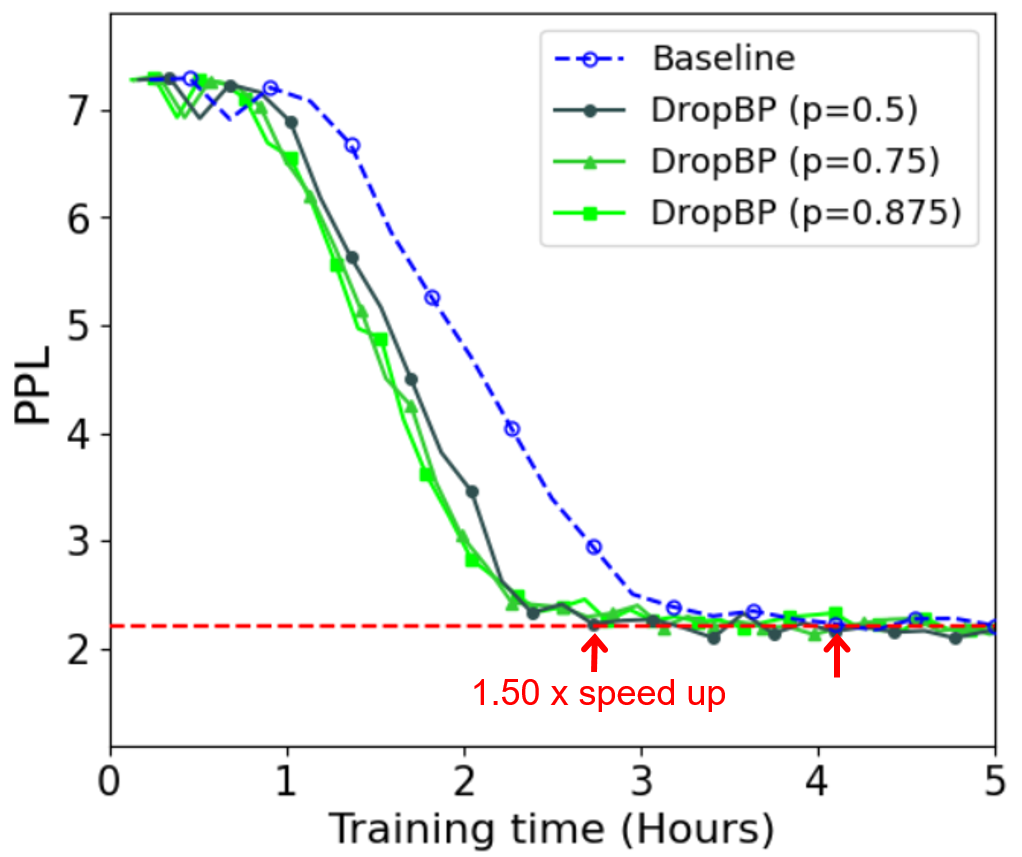}
  \caption{Validation perplexity (PPL) for fine-tuning LLaMA2-70B through QLoRA (baseline) with DropBP on the Alpaca dataset. The $p$ represents the target average drop rate for backward propagation.}
  \label{fig: 70b_qlora_loss_curve}
\end{wrapfigure}

We measured the fine-tuning time required to obtain the results in Table \ref{tb: main results}, as presented in Table \ref{tb: training time}. When using DropBP to LoRA or QLoRA, training time is reduced by 25\%, 38\%, and 44\% at drop rates of 0.5, 0.75, and 0.875, respectively. In contrast, using DropBP to Full-FT resulted in even higher training time reductions of 33\%, 50\%, and 57\% at the same drop rates. These findings align with the theoretical reduction in FLOPs due to DropBP, as detailed in Appendix \ref{appendix : training time reduction using dropbp}. We also confirmed that DropBP can significantly reduce memory usage during fine-tuning, as shown in Table \ref{tb: mt-bench}. Specifically, while not using DropBP results in a memory consumption of 57GB, applying DropBP with a drop rate of 0.875 reduces memory usage to 32GB by eliminating the storage of activation memory for dropped layers. Additionally, we evaluated the convergence speed to reach the same validation perplexity (PPL) on downstream tasks, as illustrated in Fig. \ref{fig: 70b_qlora_loss_curve} and Fig. \ref{fig: learning curve of alpaca and dolly}-\ref{fig: step curve} in Appendix \ref{appendix : learning curves when fine-tuning large language models using dropbp}. The results indicate that our DropBP increases training speed by up to 1.5$\times$ compared to the baseline in LLaMA2-70B.

\subsection{Usability of DropBP}\label{Subsection: Usability of DroBP}

In this section, we evaluate the usability of DropBP, including its ability to train on long sequence data and its training throughput in constrained environments, such as a single NVIDIA A100 GPU or Intel Gaudi2 HPU. 

\begin{table}[h!]
\caption{Available maximum sequence length for fine-tuning LLaMA2-70B using QLoRA with DropBP on a NVIDIA A100 GPU, at a micro batch size of 1.}
\centering
\footnotesize
\renewcommand{\arraystretch}{1.1}
\label{tb: max seq len}
\begin{tabular}{c|c:ccc}
\Xhline{3\arrayrulewidth}
\textbf{Method}      & \textbf{QLoRA} &                    & \textbf{w/ DropBP}   &                            \\ \hline 
\textbf{Drop Rate}   & -              & 0.5                & 0.75              & \textbf{0.875}             \\ \cdashline{1-5}
\textbf{Max Seq Len} & 0.6K           & 1.2K (2.0$\times$) & 2.0K (3.3$\times$) & \textbf{3.7K} (\textbf{6.2}$\times$) \\ \Xhline{3\arrayrulewidth}
\end{tabular}
\end{table}

We first measured the maximum sequence length that could be trained without an Out Of Memory (OOM) on a single NVIDIA A100 GPU. The results in Table \ref{tb: max seq len} indicate that our DropBP considerably increases the maximum sequence length, by up to 6.2$\times$ the baseline when the drop rate was 0.875. This is because DropBP allows skipping certain layers during backward propagation, eliminating the need to store activations required for calculating parameter gradients of those skipped layers. We believe that this property of DropBP will be particularly useful for fine-tuning LLMs with long-context data \cite{chen-longolora,xiong-long-context1}. 

\begin{figure}[h!]
\centering
\subfloat[Throughput in a single A100 GPU.\label{fig: A100 Throughput}]{\includegraphics[width=.45\textwidth]{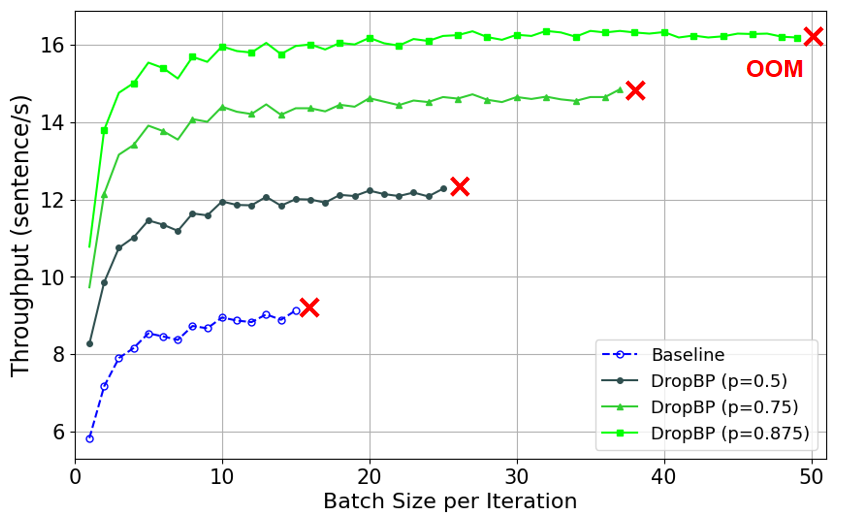}} \quad
\subfloat[Throughput in a single Gaudi2 HPU.\label{fig: Gaudi Throughput}]{\includegraphics[width=.45\textwidth]{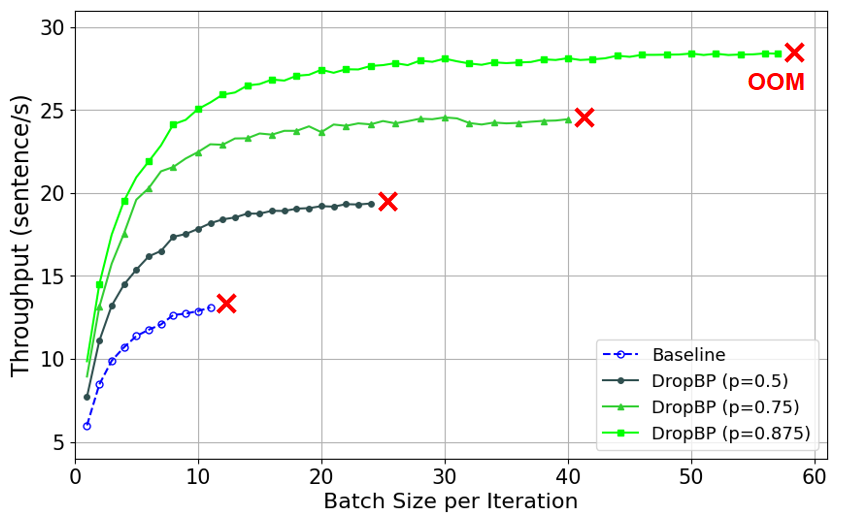}}
\caption{Throguhput (sentences/s) on a single NVIDIA A100 GPU on a single NVIDIA A100 GPU and Intel Gaudi2 HPU when fine-tuning LLaMA3-8B with a sequence length of 512.}
\label{fig: throughput}
\end{figure}

We also evaluated training throughput when full fine-tuning LLaMA3-8B using BF16 precision on a single NVIDIA A100 GPU and an Intel Gaudi2 HPU, increasing the batch size up to the point of OOM errors. As shown in Fig. \ref{fig: throughput}, applying DropBP allows for an increase in batch size per iteration by up to 3.3$\times$ on the NVIDIA A100 GPU and 5.2$\times$ on the Intel Gaudi2 HPU, ensuring high hardware utilization and scalability. Furthermore, DropBP demonstrates a sustained increase in throughput over the baseline at an identical batch size. Ultimately, with a drop rate of 0.875, DropBP achieves a throughput of 16.4 sentences/s on the NVIDIA A100 GPU and 28.4 sentences/s on the Intel Gaudi2 HPU, increasing by 79\% and 117\% over the baseline, respectively.  

\subsection{Ablation Study}\label{Subsection: Ablation Study}

\paragraph{Impact of the Number of Submodules}

\begin{wrapfigure}{!r}{0.46\textwidth}
  \centering
  \includegraphics[width=1.0\linewidth]{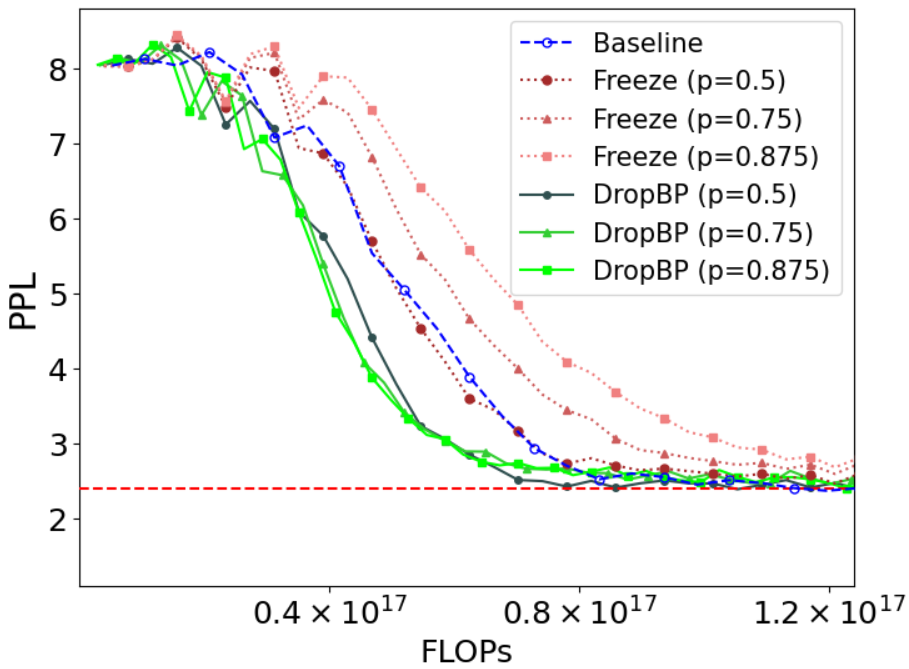}
  \caption{Validation perplexity (PPL) for fine-tuning LLaMA2-7B through LoRA with layer freezing or DropBP on the Alpaca dataset.}
  \label{fig: freezing vs dropbp}
\end{wrapfigure}

We conducted an ablation study to investigate the impact of the number of trainable submodules on the fine-tuning of LLMs. This study compared DropBP, which trains varying submodules randomly at each iteration, with layer freezing, which trains submodules composed of only upper layers. Here, the skip rate \( p \) denotes the drop rate in DropBP and the proportion of layers that are frozen in the layer freezing.

First, we analyzed the number of submodules trained by layer freezing and DropBP. In the case of layer freezing, the lower \( 2np \) layers are frozen and only the remaining \( 2n(1-p) \) layers are trained. In this case, the number of trainable upper submodules is \( 2^{2n(1-p)} \). In contrast, DropBP randomly drops \( 2np \) layers at each iteration, allowing it to train all submodules with a depth of \( 2n(1-p) \) or less without the restriction of training only the submodules composed of the upper layers. In this scenario, since the number of different submodules at depth \( i \) in the entire network is \( _{2n}C_{i} \), DropBP can train \( \sum_{i=0}^{2n(1-p)}\) \(_{2n}C_{i} \) submodules.

As shown in Table \ref{tb: number of submodules}, when fine-tuning LLaMA2-7B using layer freezing or DropBP with a high skip rate of 0.875, we observed a significant 1.8\% decrease in accuracy with layer freezing compared to the baseline, while DropBP exhibited a relatively smaller accuracy decrease of 1.0\%. Furthermore, as illustrated in Fig. \ref{fig: freezing vs dropbp}, the convergence speed to the same validation PPL on the downstream task is much slower for layer freezing compared to DropBP, especially at high skip rates, where it converges even more slowly than the baseline. We believe this is due to the ability of DropBP to train a relatively larger number of submodules (\( \sum_{i=0}^{8} \) \(_{64}C_{i} \)), compared to the fewer submodules trained by layer freezing (\( 2^{8} \)). Moreover, when fine-tuning LLaMA2-70B, DropBP resulted in a 0.5\% increase in MMLU 5-shot accuracy compared to the baseline, despite a high skip rate of 0.875. This improvement is due to the large number of layers in LLaMA2-70B, enabling DropBP to train deeper and more numerous submodules (\( \sum_{i=0}^{20} \) \(_{160}C_{i} \)) even with a high skip rate of 0.875.


\begin{table}[h!]
\caption{The number of submodules being trained and test accuracy on the 5-shot MMLU tasks with layer freezing or DropBP on the Alpaca datasets.}
\centering
\footnotesize
\renewcommand{\arraystretch}{1.3}
\label{tb: number of submodules}
\begin{tabular}{c|ccc|cc}
\Xhline{3\arrayrulewidth}
                         & \multicolumn{3}{c|}{\textbf{LLaMA2-7B}}                                               & \multicolumn{2}{c}{\textbf{LLaMA2-70B}}                             \\ \hline
\textbf{Method}          & \multicolumn{1}{c:}{\textbf{LoRA}} & \textbf{LoRA+Freeze}      & \textbf{LoRA+DropBP} & \multicolumn{1}{c:}{\textbf{QLoRA}} & \textbf{QLoRA+DropBP}         \\ \hline
\textbf{Drop Rate}       & \multicolumn{1}{c:}{-}             & 0.875                     & 0.875                & \multicolumn{1}{c:}{-}              & 0.875                         \\ \cdashline{1-6}
\textbf{\# of Submodules} & \multicolumn{1}{c:}{$2^{64}$}      & $2^{8}$ & $\sum_{i=0}^{8}$ $_{64}C_{i}$          & \multicolumn{1}{c:}{$2^{160}$}   & $\sum_{i=0}^{20}$ $_{160}C_{i}$  \\
\textbf{Accuracy (\%)}        & \multicolumn{1}{c:}{44.7}          & 42.9 (-1.8)               & 43.7 (-1.0)          & \multicolumn{1}{c:}{68.3}           & 68.8 (+0.5)                   \\ \Xhline{3\arrayrulewidth}
\end{tabular}
\end{table}

\paragraph{Impact of Sensitivity-based Drop Rate}

\begin{wraptable}{r}{0.5\textwidth}
\caption{Test accuracy on the 0-shot commonsense reasoning tasks when fine-tuning LLaMA2-7B and 13B through LoRA with DropBP at uniform or sensitivity-based drop rate on the Alpaca datasets. The target average drop rate is 0.875.}
\centering
\small
\label{tb: sensitivity 1e-4 3e-4}
\resizebox{0.5\textwidth}{!}{%
\begin{tabular}{ccccc}
\Xhline{2\arrayrulewidth}
\textbf{LLaMA2} & \multicolumn{2}{c}{\textbf{7B}} & \multicolumn{2}{c}{\textbf{13B}} \\ \cdashline{1-5}
\textbf{LR} & \textbf{1e-4} & \textbf{3e-4} & \textbf{1e-4} & \textbf{3e-4} \\ \Xhline{2\arrayrulewidth}
LoRA & 65.7 & 66.0 & 68.2 & 68.0 \\ \cdashline{1-5}[1pt/1pt]
+DropBP (uniform) & 66.4 & 63.1 & 66.6 & 65.8 \\
+DropBP (sens) & 66.6 & 64.7 & 67.7 & 67.3 \\ \Xhline{2\arrayrulewidth}
\end{tabular}}
\end{wraptable}

\begin{figure}[b!]
\centering
\subfloat[Distribution of drop rates determined by sensitivity when the average drop rate is set to 0.875.\label{fig: sensitivity for ablation stduy}]{\includegraphics[width=.55\textwidth]{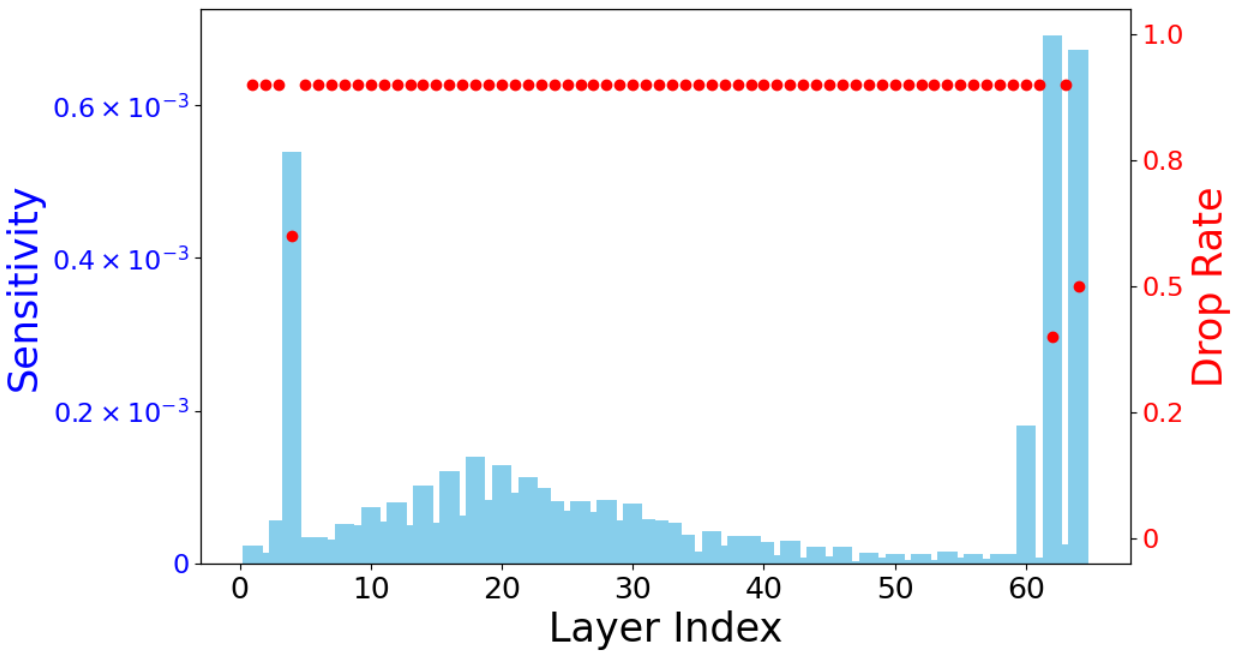}} \quad
\subfloat[Validation PPL with uniform and sensitivity-based allocated drop rates.\label{fig: loss curve for ablation stduy}]{\includegraphics[width=.38\textwidth]{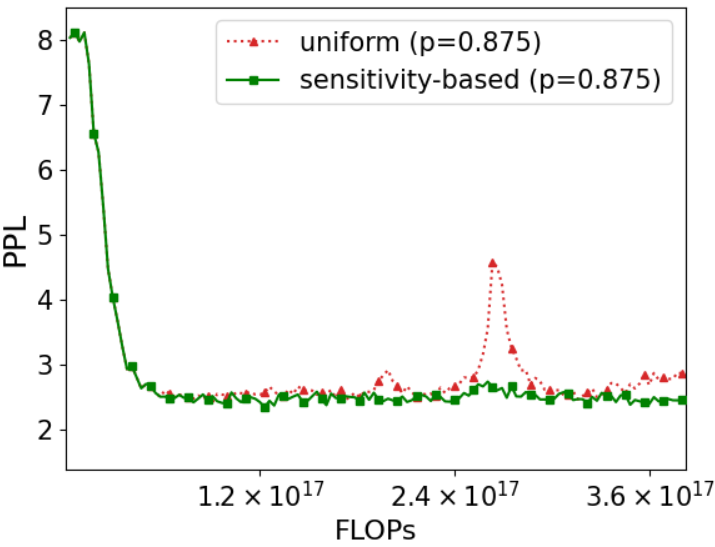}}
\caption{Distribution of drop rates and the validation PPL when fine-tuning LLaMA2-7B through LoRA with DropBP at uniform or sensitivity-based drop rate on Alpaca datasets.}
\label{fig: ablation stduy}
\end{figure}

We also conducted an ablation study to analyze the effectiveness of sensitivity-based drop rate allocations. First, we identified the sensitivity of different layers by calculating Eq. \ref{eq: sensitivity calculation} during the training of LLMs in various scenarios, as illustrated in Fig. \ref{fig: sensitivity for ablation stduy} and Fig. \ref{fig: distribution of drop rates determined by sensitivity} in Appendix \ref{appendix: distribution of sensitivity-determined drop rates}. While the distribution varies slightly depending on the number of parameters, fine-tuning approach, and target average drop rate, there is a consistent tendency to assign importance to both the initial and final layers. Consequently, drop rates for these layers are allocated to be lower by a simple greedy algorithm, as explained in Section \ref{Subsection: Implementation}

Additionally, we fine-tuned the LLaMA2-7B and 13B using DropBP on Alpaca datasets, comparing sensitivity-based allocated drop rates with uniform drop rate. In detail, we compared the average accuracy of commonsense reasoning tasks when fine-tuning the models with a learning rate of 1e-4 and 3e-4, as shown in Table \ref{tb: sensitivity 1e-4 3e-4}. Note that the PPL for fine-tuning LLaMA2-7B in Fig. \ref{fig: loss curve for ablation stduy} corresponds to a learning rate of 3e-4. The results indicate that sensitivity-based drop rates achieved a 1.6\% higher accuracy compared to uniform drop rates with a relatively high learning rate of 3e-4, while there was no significant difference in accuracy when the learning rate was set to 1e-4 in LLaMA2-7B. Fig. \ref{fig: loss curve for ablation stduy} also shows that sensitivity-based drop rates consistently stabilized the convergence of validation loss, whereas uniform drop rates occasionally diverged when the learning rate was set to 3e-4 in LLaMA2-7B. This phenomenon is even more pronounced with LLaMA2-13B, resulting in a 1.1\% increase in accuracy through sensitivity-based drop rate allocation, even with a low learning rate of 1e-4. In other words, sensitivity-based drop rate allocation helps stabilize the training process, especially in the case of large learning rates or larger models.

\section{Related Works}\label{Section: related works}
\paragraph{Parameter-efficient fine-tuning}
When fine-tuning LLM, substantial amount of memory is required to store parameters, gradients, and optimizer states. LoRA \cite{Hu-LoRA} successfully reduces the memory allocated to gradients and optimizer states by inserting trainable rank decomposition matrices into the linear layers of the model while keeping the original LLM parameters frozen. LLaMA-Adapter \cite{Zhang-LLaMA-Adapter} and LLaMA-Adapter V2 \cite{Gao-LLaMA-ADAPTER-V2} significantly reduce training memory using trainable adoption prompts and zero-initialized attention mechanisms. Some studies attempt to reduce not only the memory footprint of gradients and optimizer states but also that of parameters by considering quantization. PEQA \cite{Kim-PEQA}, for instance, quantizes the original LLM parameters into a low-bit format and fine-tunes only the scale factor, thus saving memory for parameters during training. QLoRA \cite{Dettmers-QLoRA} and QA-LoRA \cite{Xu-QALoRA}, built upon LoRA, also employ quantization on the original LM parameters, significantly reducing parameter memory during training. Our DropBP is orthogonal and easily combinable with these PEFT techniques, enabling memory and computationally efficient fine-tuning.

\paragraph{Parallelism}
Parallelism techniques are widely used to accelerate training LLM using multiple GPU efficiently. Data parallelism \cite{Li-DDP} is a technique that involves dividing data along the batch dimension for training across multiple GPUs, which still requires sufficient memory to load the entire model on each GPU. Conversely, tensor parallelism \cite{Shoeybi-MegatronLM, Smith-megtron-lm-530b, Kofghikzngi-megtron-sequence-2} partitions the model across GPUs, dividing matrix multiplication operations column-wise and row-wise. Pipeline parallelism \cite{Huang-GPIPE, Harlap-PipeDream, BPIPE-Tae} involves partitioning the model depth-wise across GPUs, which enables efficient pipeline scheduling. The Zero Redundancy Optimizer (ZeRO) \cite{ZeRO-Samyam} and Fully Sharded Data Parallelism (FSDP) \cite{Zhao-FSDP} shard parameters, gradients, and optimizer states across multiple GPUs, retrieving parameters when needed to restore their non-partitioned form, enabling the overlapping of computation and communication during training. While these parallelism techniques are designed to efficiently manage the massive computational costs across multiple GPUs, our DropBP specifically aims to reduce the inherent computational costs required for training process.

\paragraph{Layer dropping}
Stochastic Depth \cite{Huang-StochasticDepth}, the first approach to randomly drop layers during neural network training, reduces overfitting and costs in image recognition. Layerdrop \cite{Fan-layerdrop} randomly drops layers during training and selectively uses some layers during inference, accelerating both processes for transformers. Progressive Layer Dropping (PLD) \cite{Zhang-PLD} progressively increases the drop rate across depth and iterations, improving training speed without accuracy degradation in transformers. These techniques speed up pretraining of small transformer models like BERT \cite{Devlin-BERT} by dropping layers during the entire training process, whereas DropBP, specific to fine-tuning LLMs, drops layers only during backward propagation. Consequently, as detailed in Appendix \ref{appendix: comparisons between layer dropping}, our DropBP achieves higher performance compared to these layer dropping when fine-tuning LLMs.

\section{Conclusion}
We propose DropBP, an effective algorithm that accelerates the fine-tuning of LLMs by randomly dropping layers during backward propagation, which can be orthogonally integrated into both full-fine tuning and parameter-efficient fine-tuning. We developed the DropBP library as a user-friendly PyTorch extension to facilitate easy integration into existing training codes. Experimental results demonstrate that DropBP significantly accelerates training speed during the fine-tuning of LLMs, achieving comparable accuracy to baseline fine-tuning. Furthermore, DropBP reduces activation memory, enabling long-context training and increasing batch size on limited resources. Consequently, applying DropBP enables a 79\% higher throughput on an NVIDIA A100 GPU and a 117\% higher throughput on an Intel Gaudi2 HPU.

\section*{Acknowledgment}

This research was supported in part by the NAVER-Intel Co-Lab. The work was conducted by Seoul National University and reviewed by both NAVER and Intel.



\bibliographystyle{unsrt}
\bibliography{main}

\newpage
\appendix
\section*{Appendices}
\section{The importance of short paths in residual networks} \label{appendix: the importance of short paths in residual networks}

\begin{figure}[!ht]
\centering
\subfloat[Distribution of path length. \label{fig: distribution}]{\includegraphics[width=.33\textwidth]{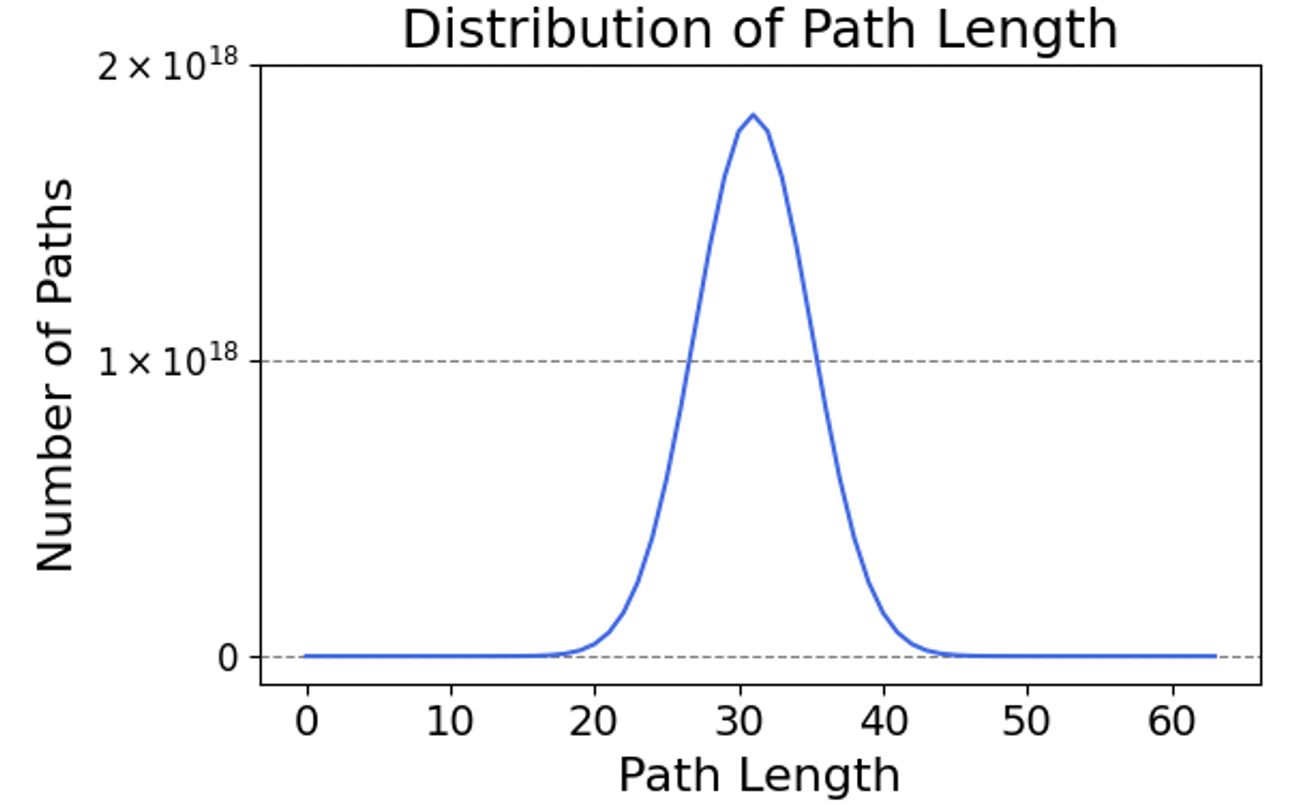}}
\subfloat[Gradient per path length. \label{fig: path grad}]{\includegraphics[width=.33\textwidth]{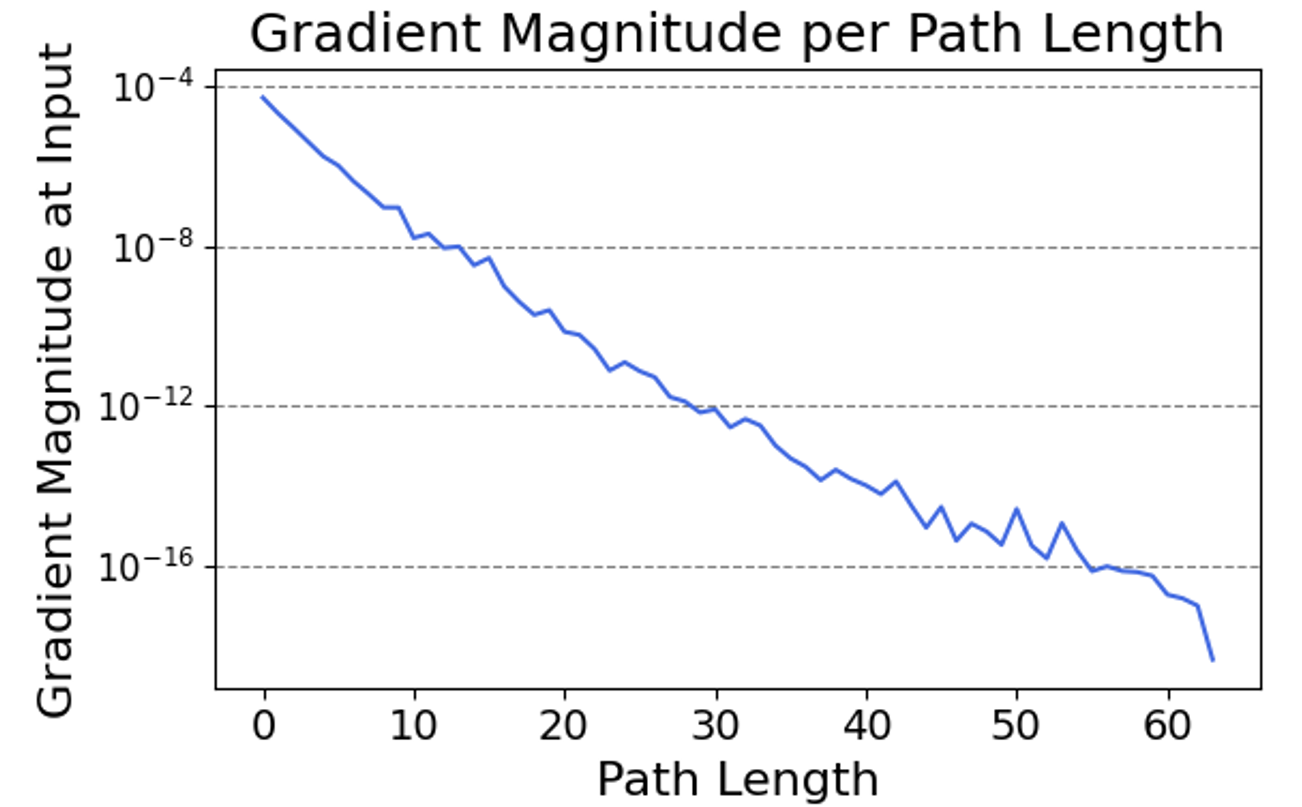}}
\subfloat[Total gradient per path length. \label{fig: total grad}]{\includegraphics[width=.33\textwidth]{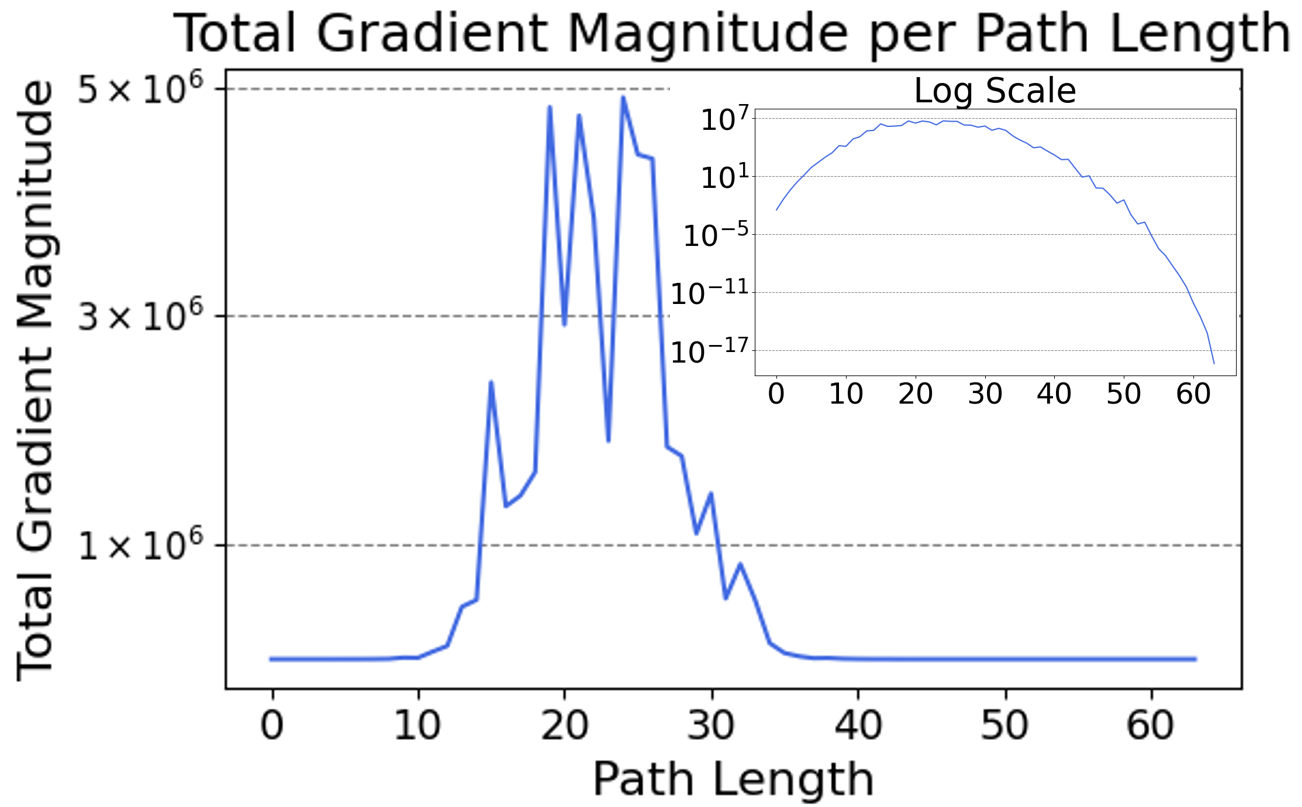}}
\caption{The impact of path length for fine-tuning LLaMA2-7B.}
\label{fig: short paths}
\end{figure}

In Section \ref{Subsection: Dropping Backpropagation}, we interpret transformer models as a collection of numerous blocks, each composed of various modules with residual connections. Our hypothesis is that we can fine-tune LLMs well by training only certain shallow submodules. To theoretically analyze this hypothesis, we measured the impact of submodules based on their path lengths in LLaMA2-7B, as shown in Fig. \ref{fig: short paths}. Specifically, we followed these steps:

\begin{itemize}
 \item We first perform a forward pass through the entire network.
 \item During the backward pass, we randomly sample $k$ residual blocks, which are back-propagated without passing through skip connections, while the remaining $n-k$ blocks are bypassed through the skip connections.
\item We then measure the norm of the gradient at the input.
\end{itemize}

We take 100 measurements for each path length $k$. Subsequently, we multiply by the distribution of all possible path lengths, which follows a Binomial distribution, to quantify the gradient contribution from paths of a specific length.

In Fig. \ref{fig: path grad}, we observed that the gradient per path length decreases as the path length increases. Consequently, Fig. \ref{fig: total grad} demonstrates that shorter path lengths have a greater impact on the gradient in LLaMA2-7B. These observations are consistent with the existing findings \cite{Veit-shallownet} in ResNet \cite{ResNet}, which attributed this phenomenon to vanishing gradients. Therefore, our DropBP enables effective training LLMs by focusing on training important short submodules.

\section{Theoretical FLOPs and Actual Training Time Using DropBP} \label{appendix : training time reduction using dropbp}

\begin{table}[!b]
\renewcommand{\arraystretch}{1.2}
\caption{Training time (ms) per iteration for a sequence length of 512 through Full-FT, LoRA or QLoRA using DropBP. Mixed refers to mixed precision training \cite{micikevicius-AMP} using BFloat16 (BF16) and 32-bit. MBS is denoted as the micro batch size. FW, BW, and Total respectively denote the time consumed for forward propagation, backward propagation, and the entire training process.}
\centering
\small
\label{tb: training time for main results}
\resizebox{1.0\textwidth}{!}{%
\begin{tabular}{c|ccccccc}
\Xhline{3\arrayrulewidth}
\textbf{Model} & \textbf{Method} & \textbf{Precision} & \textbf{MBS}      & \textbf{Drop Rate} & \textbf{FW} & \textbf{BW} & \textbf{Total}        \\ \Xhline{3\arrayrulewidth}

\multirow{10}{*}{LLaMA2-7B} &LoRA & Mixed &2& 0  & 159  & 161   & 320                  \\ \cdashline{2-8}[1pt/1pt]
&LoRA+DropBP & Mixed &2& 0.5              & 159  & 81 (-50\%)   & 239 (-25\%)          \\
&LoRA+DropBP & Mixed &2& 0.75             & 159  & 43 (-74\%)  & 201 (-37\%)          \\
&\textbf{LoRA+DropBP} & Mixed &2& 0.875   & 158  & 23 (-86\%)   & \textbf{181 (-43\%)} \\ \cdashline{2-8}[1pt/1pt]
&Full-FT & BF16 &2& 0  & 91  & 192   & 283                 \\ \cdashline{2-8}[1pt/1pt]
&Full-FT+DropBP & BF16 &2& 0.5              & 91  & 98 (-49\%)   & 189 (-33\%)          \\
&Full-FT+DropBP & BF16 &2& 0.75             & 91  & 52 (-73\%)  & 143 (-50\%)          \\
&\textbf{Full-FT+DropBP} & BF16 &2& 0.875   & 91  & 30 (-85\%)   & \textbf{121 (-57\%)} \\ \Xhline{1\arrayrulewidth}
\multirow{5}{*}{LLaMA2-13B} &LoRA & BF16 &2& 0   & 186   & 236   & 423                  \\ \cdashline{2-8}[1pt/1pt]
&LoRA+DropBP & BF16 &2& 0.5              & 186  & 119 (-50\%)   & 306 (-28\%)          \\
&LoRA+DropBP & BF16 &2& 0.75             & 187  & 64 (-73\%)  & 251 (-41\%)          \\
&\textbf{LoRA+DropBP} & BF16 &2& 0.875   & 187  & 33 (-86\%)  & \textbf{219 (-48\%)} \\ \Xhline{1\arrayrulewidth}


\multirow{5}{*}{LLaMA2-70B} &QLoRA & BF16 &1& 0 & 1033         & 1100         & 2133                  \\ \cdashline{2-8}[1pt/1pt]
&QLoRA+DropBP &BF16 &1 & 0.5              & 1034  & 566 (-49\%)  & 1599 (-25\%)          \\
&QLoRA+DropBP &BF16 &1 & 0.75             & 1033  & 290 (-74\%)  & 1323 (-38\%)          \\
&\textbf{QLoRA+DropBP} &BF16 &1 & 0.875   & 1032  & 158 (-86\%)  & \textbf{1191 (-44\%)} \\ \Xhline{3\arrayrulewidth}
\end{tabular}
}
\end{table}

In this section, we calculate the theoretical FLOPs reduction afforded by DropBP and compare this reduction to the actual training time reduction as shown in Table \ref{tb: training time for main results}. As outlined in Section \ref{Section: Background}, the computational costs arise from output activation calculations by Eq. \ref{eq: forward propagation} during forward propagation, and input and parameter gradient calculations  by Eqs. \ref{eq: backward propagation} and \ref{eq:parameter updates} during backward propagation. We denote the FLOPs for these operations as $F_{out}$, $F_{grad}$, and $F_{param}$, respectively. Therefore, the total FLOPs for the backpropagation algorithm can be calculated by the following equation:

\begin{align}
F_{T} &= F_{fw} + F_{bw}\notag \\
&= F_{out} + F_{grad} + F_{param}
\label{eq: total flops of bp}
\end{align}

where $F_{T}$ represents the FLOPs during the entire training process, $F_{fw}$ for forward propagation (i.e. $F_{fw}=F_{out}$), and $F_{bw}$ for backward propagation  (i.e. $F_{bw}=F_{grad}+F_{param}$). DropBP reduces FLOPs for backward propagation by a target average drop rate ($p_{avg}$). Therefore, total FLOPs in DropBP can be formulated as below:

\begin{align}
F_{T} &= F_{fw} + (1-p_{avg})F_{bw}\notag\\
&= F_{out} + (1-p_{avg})(F_{grad} + F_{param})
\label{eq: total flops of DropBP}
\end{align}

Consequently, the theoretical FLOPs reduction ratio by DropBP can be represented as follow:

\begin{align}
\text{Reduction Ratio by DropBP:} \quad \frac{p_{avg}(F_{grad} + F_{param})}{F_{out} + F_{grad} + F_{param}}
\label{eq: reduction ratio}
\end{align}

Note that in full fine-tuning (Full-FT), the computational costs for output calculations, input gradient calculations, and parameter gradient calculations are nearly identical (i.e., $F_{out}=F_{grad}=F_{param}$). Conversely, in parameter-efficient fine-tuning techniques (PEFT) such as LoRA and QLoRA, the costs of calculating parameter gradients are negligible ($F_{out}=F_{grad}, \,\, F_{param}=0$) due to a very small number of trainable parameters and the freezing of original LLM parameters. By substituting this into Eq. \ref{eq: reduction ratio}, the theoretical FLOPs reduction ratio by DropBP can be expressed as:

\begin{gather}
\text{Reduction ratio in Full-FT:} \quad \frac{2}{3}p_{avg} \label{eq: Full-FT reduction ratio} \\ 
\text{Reduction Ratio in PEFT:} \quad \frac{1}{2}p_{avg} \label{eq: LoRA reduction ratio}
\end{gather}

Therefore, with target average drop rates of 0.5, 0.75, and 0.875, DropBP achieves theoretical FLOPs reductions in Full-FT of 33\%, 50\%, and 58\%, respectively, according to Eq. \ref{eq: Full-FT reduction ratio}. This aligns with the actual training time reduction when utilizing DropBP in Full-FT as shown in Table \ref{tb: training time} and Table \ref{tb: training time for main results}. This trend is also evident when utilizing DropBP in LoRA and QLoRA. According to Eq. \ref{eq: LoRA reduction ratio}, the reductions in FLOPs for various target average drop rates of 0.5, 0.75, 0.875 are derived as 25\%, 38\%, and 44\%, respectively. This closely aligns with the actual training time reductions observed when DropBP is applied to LoRA and QLoRA as demonstrated in Table \ref{tb: training time for main results}.

\newpage

\section{Convergence Speed Up Using DropBP} \label{appendix : learning curves when fine-tuning large language models using dropbp}

\begin{figure}[!ht]
\centering
\subfloat[LLaMA2-7B w/ LoRA (Alpaca)]{\includegraphics[width=.36\textwidth]{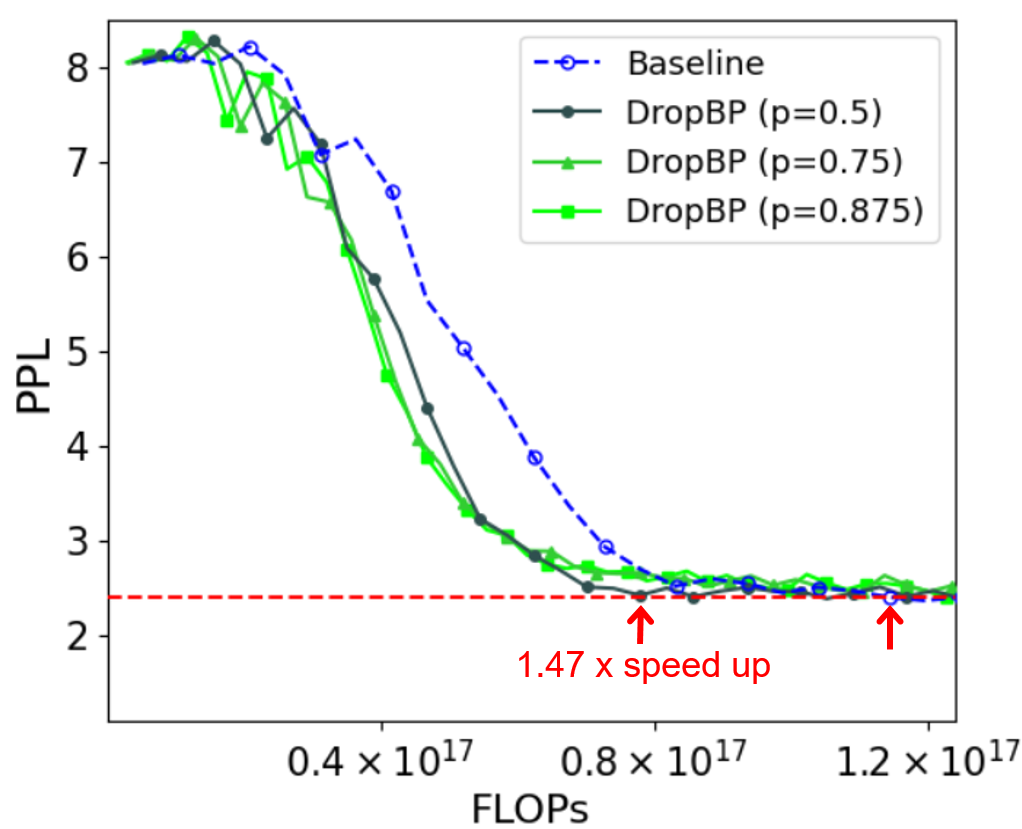}}\qquad 
\subfloat[LLaMA2-7B w/ LoRA (Dolly)]{\includegraphics[width=.36\textwidth]{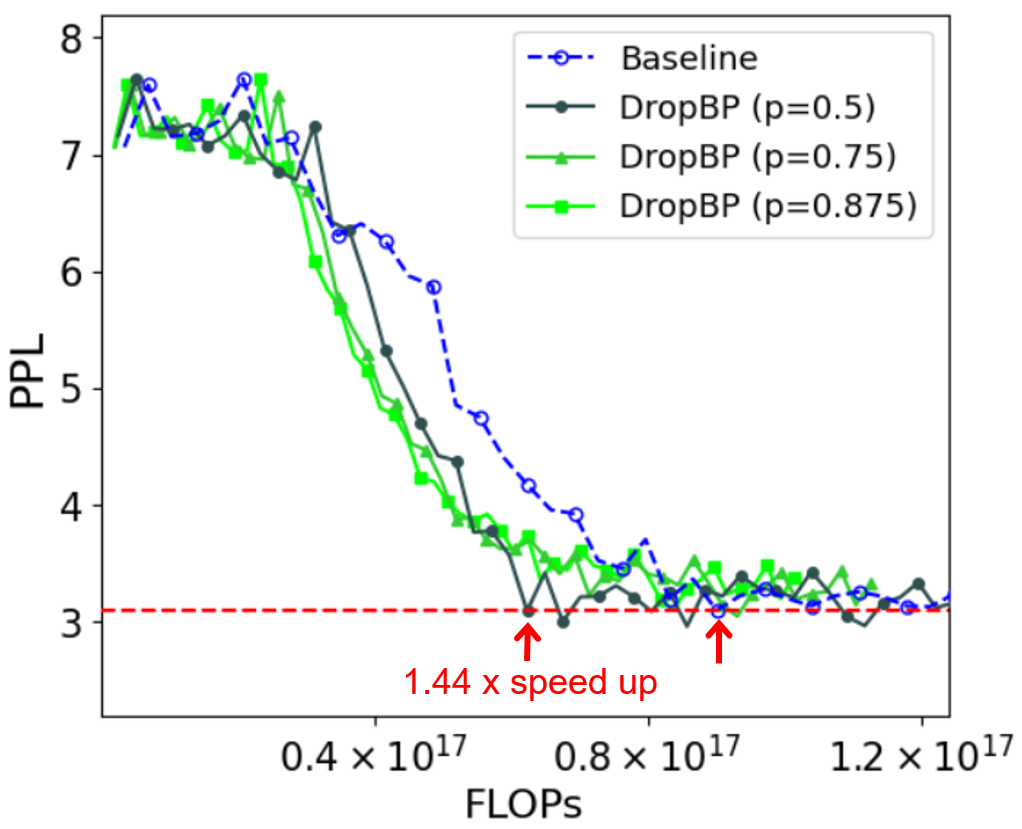}}\\

\subfloat[LLaMA2-7B w/ Full-FT (Alpaca)\label{fig: loss curve for Full-FT-Alpaca}]{\includegraphics[width=.36\textwidth]{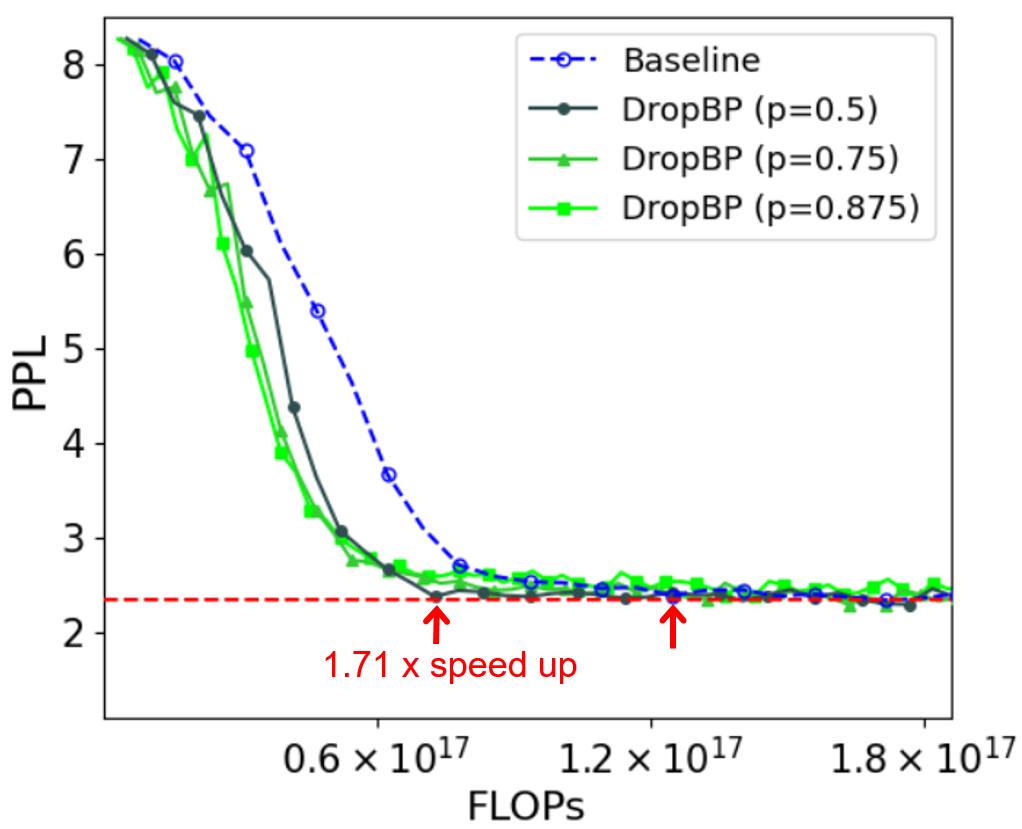}}\qquad 
\subfloat[LLaMA2-7B w/ Full-FT (Dolly)\label{fig: loss curve for Full-FT-Dolly}]{\includegraphics[width=.36\textwidth]{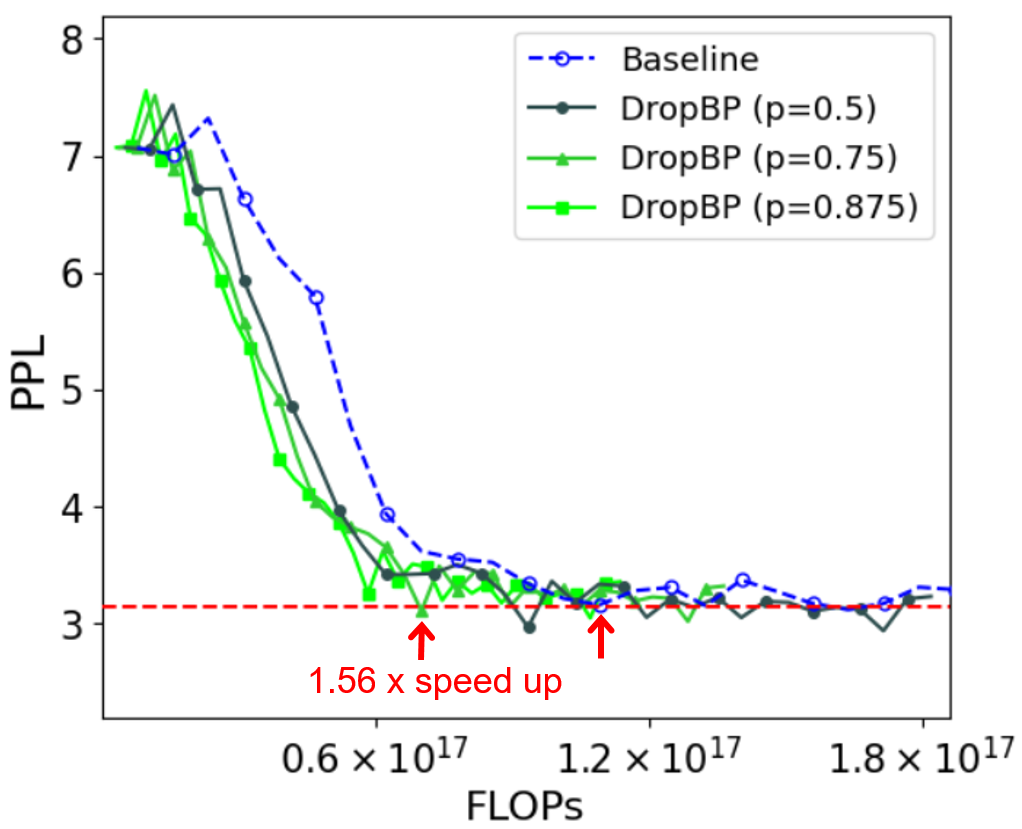}}\\

\subfloat[LLaMA2-13B w/ LoRA (Alpaca)]{\includegraphics[width=.36\textwidth]{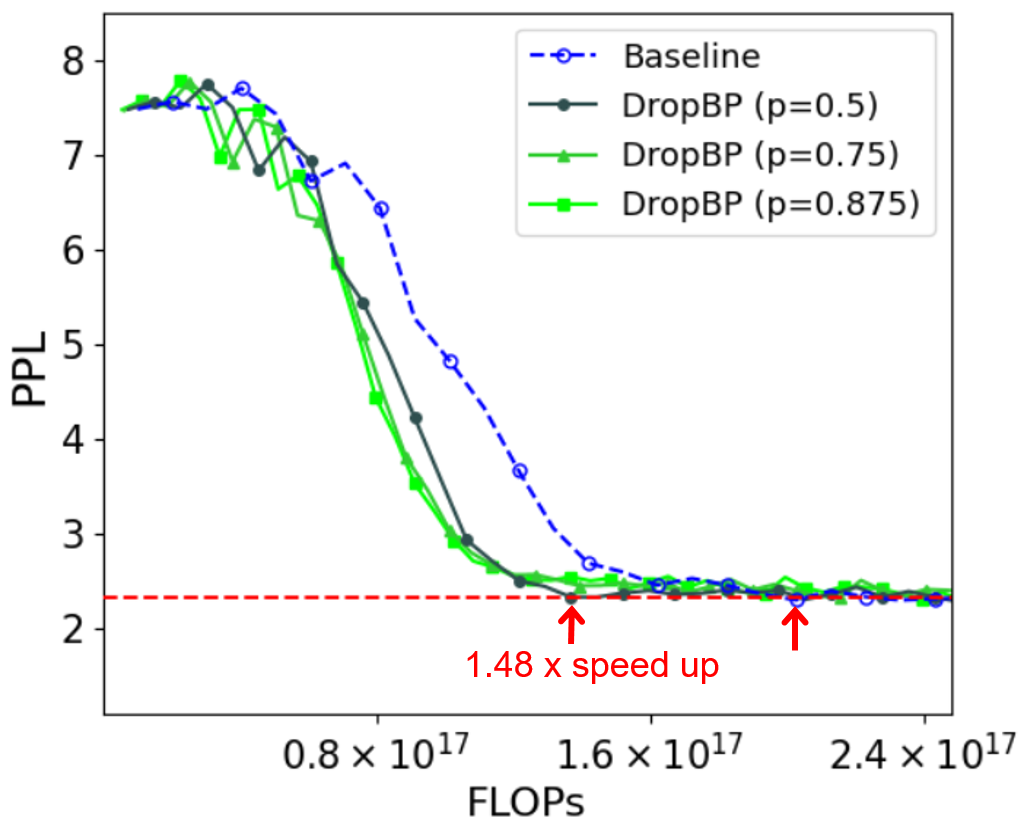}}\qquad 
\subfloat[LLaMA2-13B w/ LoRA (Dolly)]{\includegraphics[width=.36\textwidth]{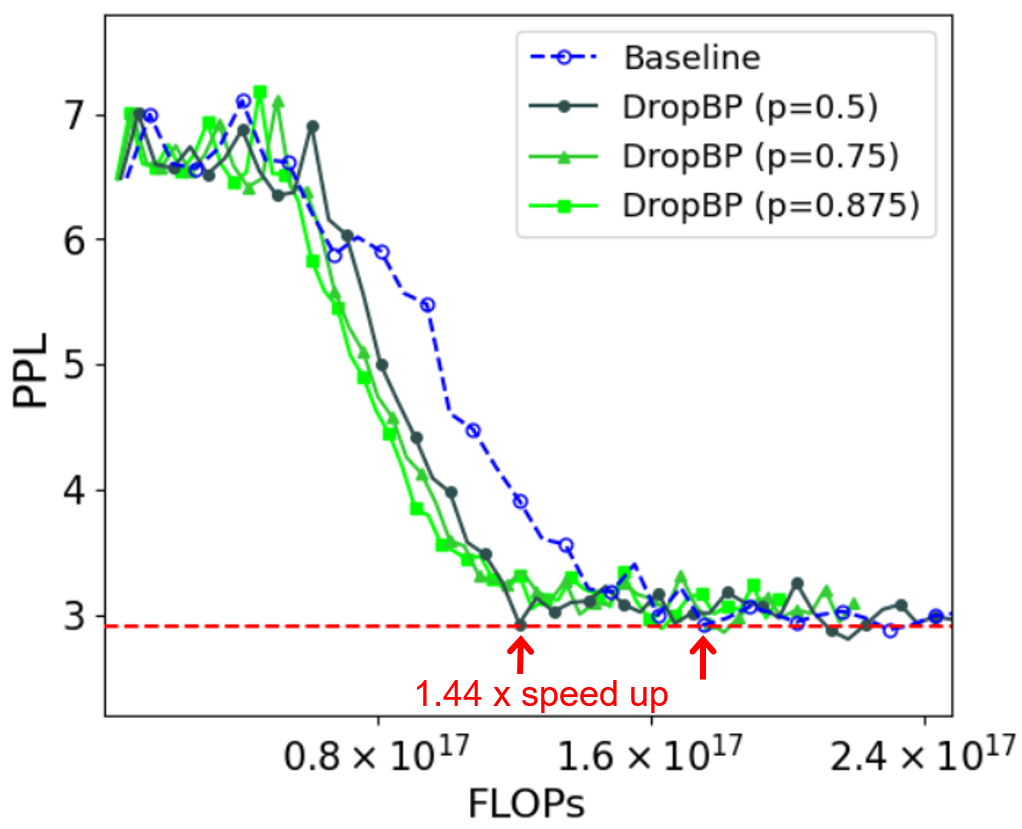}}\\

\subfloat[LLaMA2-70B w/ QLoRA (Alpaca)]{\includegraphics[width=.36\textwidth]{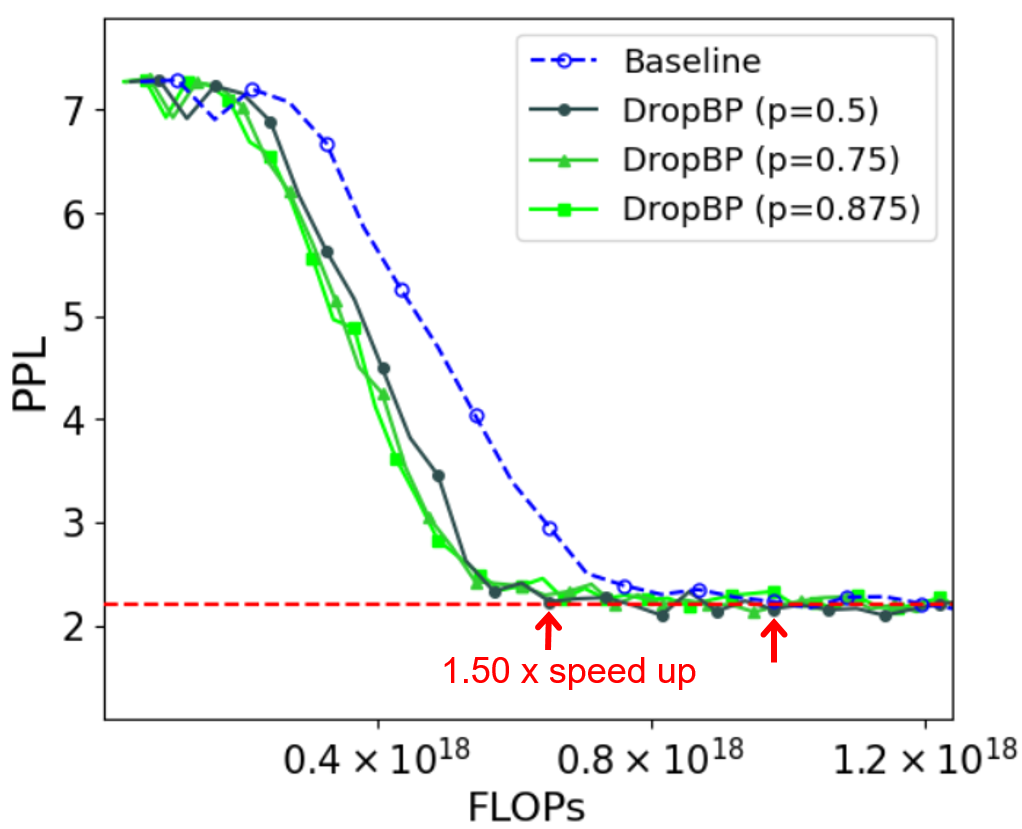}}\qquad 
\subfloat[LLaMA2-70B w/ QLoRA (Dolly)]{\includegraphics[width=.36\textwidth]{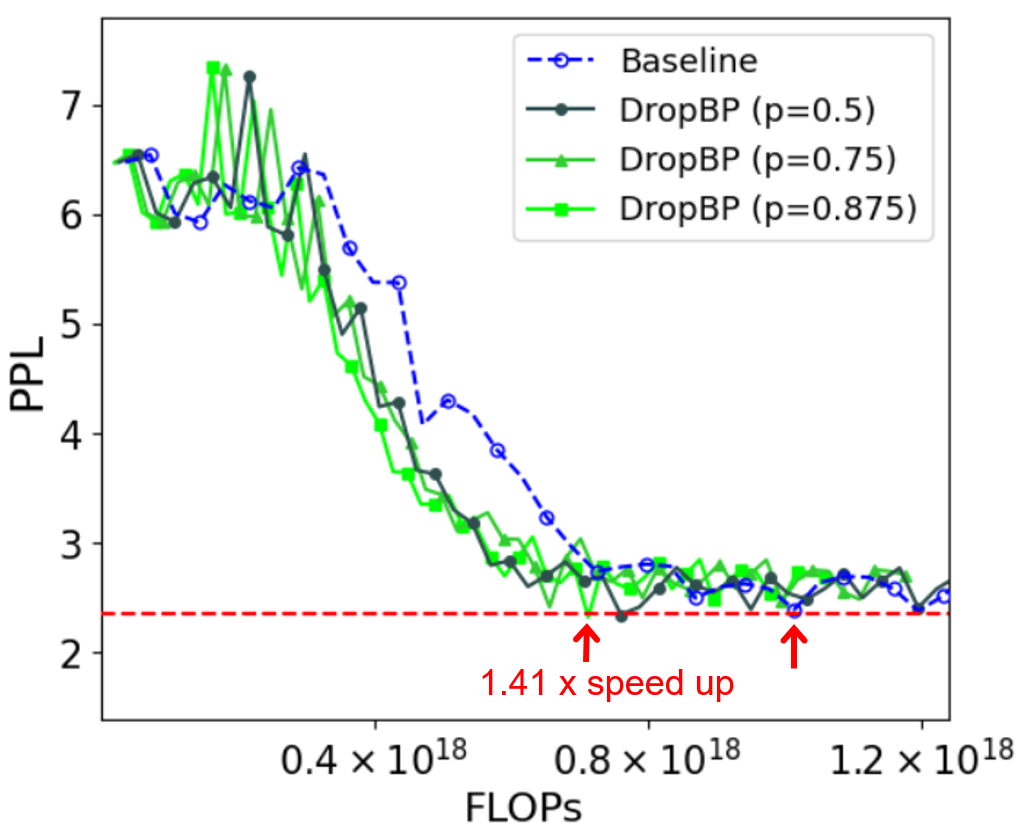}}
\caption{Validation perplexity (PPL) when fine-tuning LLaMA2 models through Full-FT, LoRA, or QLoRA using DropBP on the Alpaca and Dolly datasets.}
\label{fig: learning curve of alpaca and dolly}
\end{figure}

\newpage

\begin{figure}[!ht]
\centering
\subfloat[Perplexity curve across training steps. \label{fig: steps}]{\includegraphics[width=.4\textwidth]{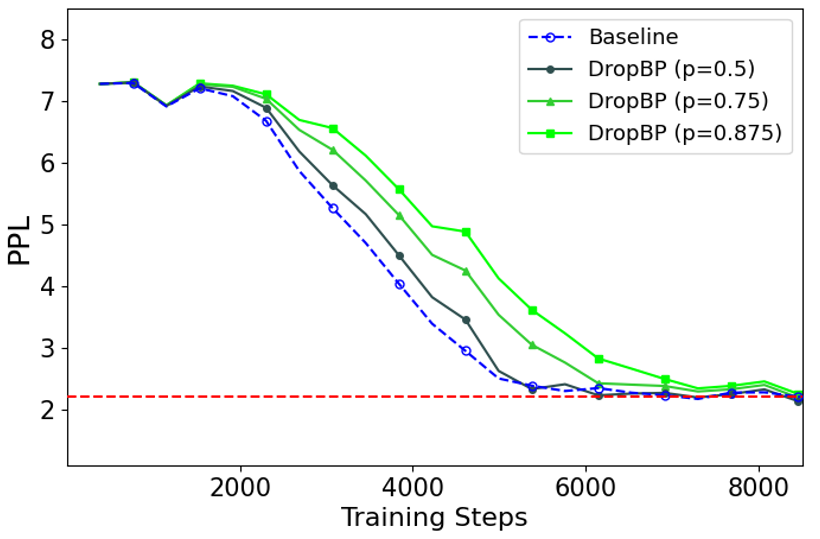}}\qquad
\subfloat[Perplexity curve across training time. \label{fig: time}]{\includegraphics[width=.4\textwidth]{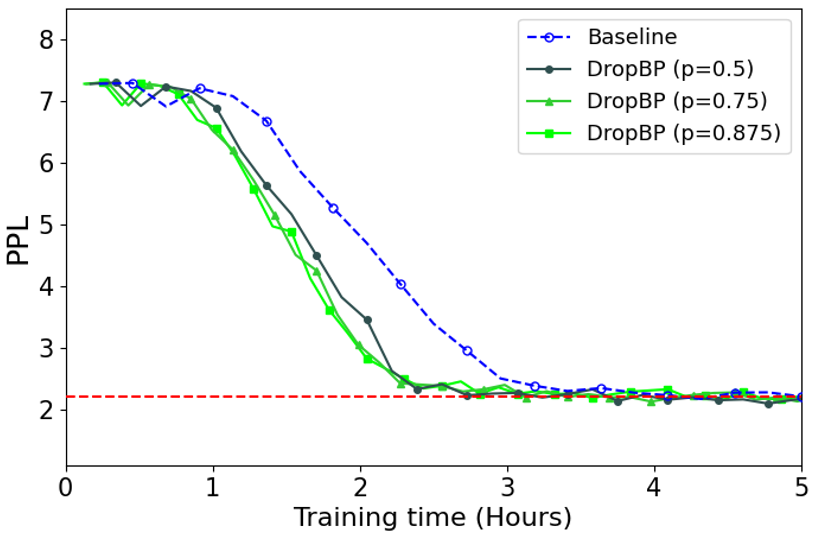}}
\caption{Training curves across training steps and time for fine-tuning LLaMA2-70B through QLoRA with DropBP on the Alpaca datasets.}
\label{fig: step curve}
\end{figure}

When analyzing training curves across training steps in Fig. \ref{fig: steps}, the convergence of loss per step at a drop rate of 0.5 is almost identical to the baseline. However, with drop rates of 0.75 and 0.875, the convergence speed per step is slower compared to baseline. Nonetheless, DropBP significantly reduces the time consumed per training step, because it skips the backward propagation computations for the dropped layers. Consequently, the convergence speed per training time is actually faster for DropBP compared to the baseline as shown in Fig. \ref{fig: time}.


\section{Distribution of Drop Rates Determined by Sensitivity}\label{appendix: distribution of sensitivity-determined drop rates}

\setcounter{figure}{7}
\begin{figure}[H]
\centering
\subfloat[LLaMA2-7B w/ LoRA + DropBP (p=0.5)]{\includegraphics[width=.45\textwidth]{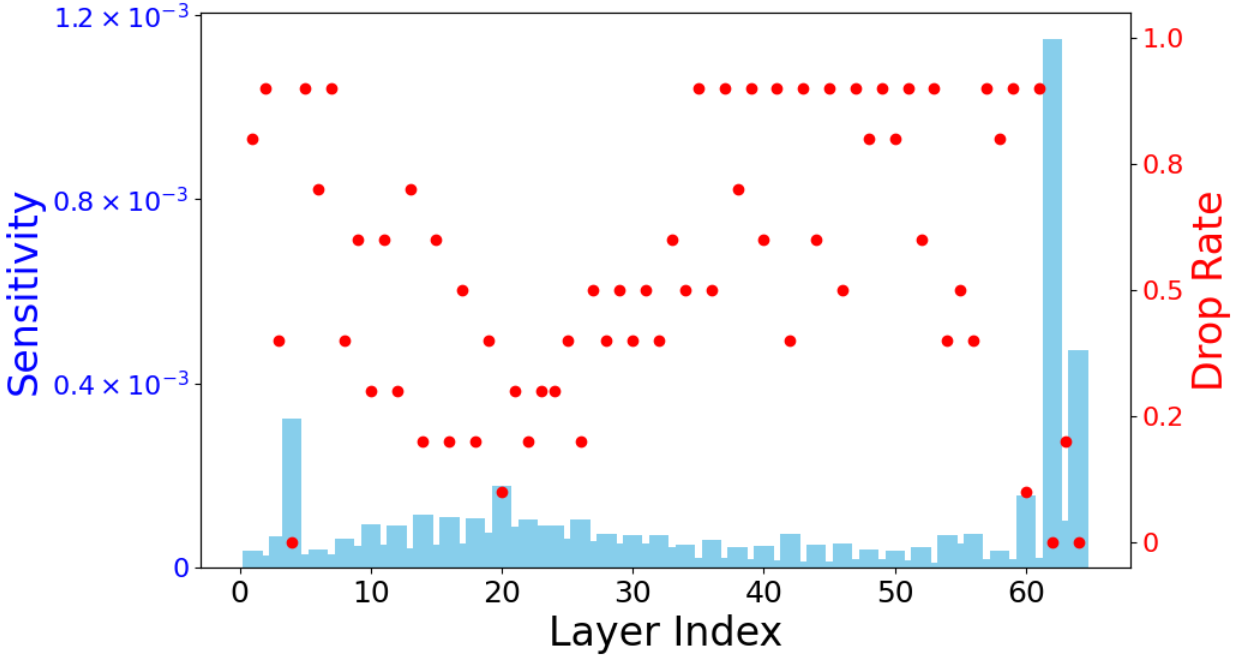}}\qquad 
\subfloat[LLaMA2-7B w/ LoRA + DropBP (p=0.875)]{\includegraphics[width=.45\textwidth]{Figure/sensitivity/alpaca/7b_0.875.png}}\\
\subfloat[LLaMA2-7B w/ Full-FT + DropBP (p=0.5)]{\includegraphics[width=.45\textwidth]{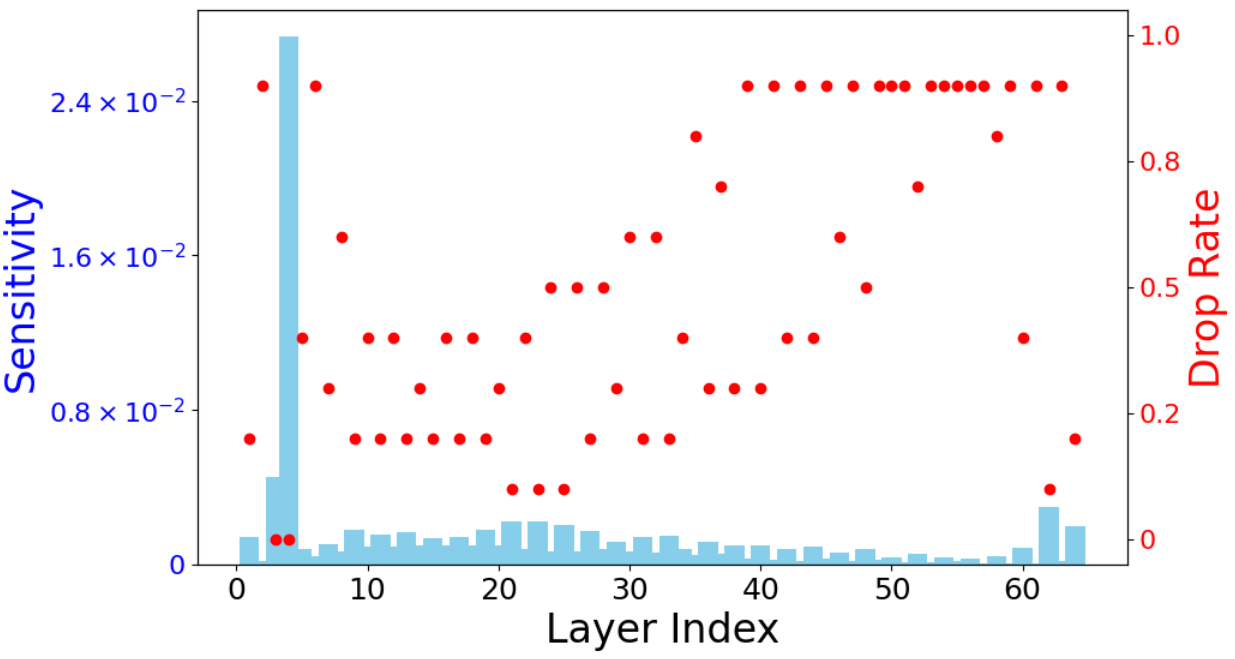}}\qquad 
\subfloat[LLaMA2-7B w/ Full-FT + DropBP (p=0.875)]{\includegraphics[width=.45\textwidth]{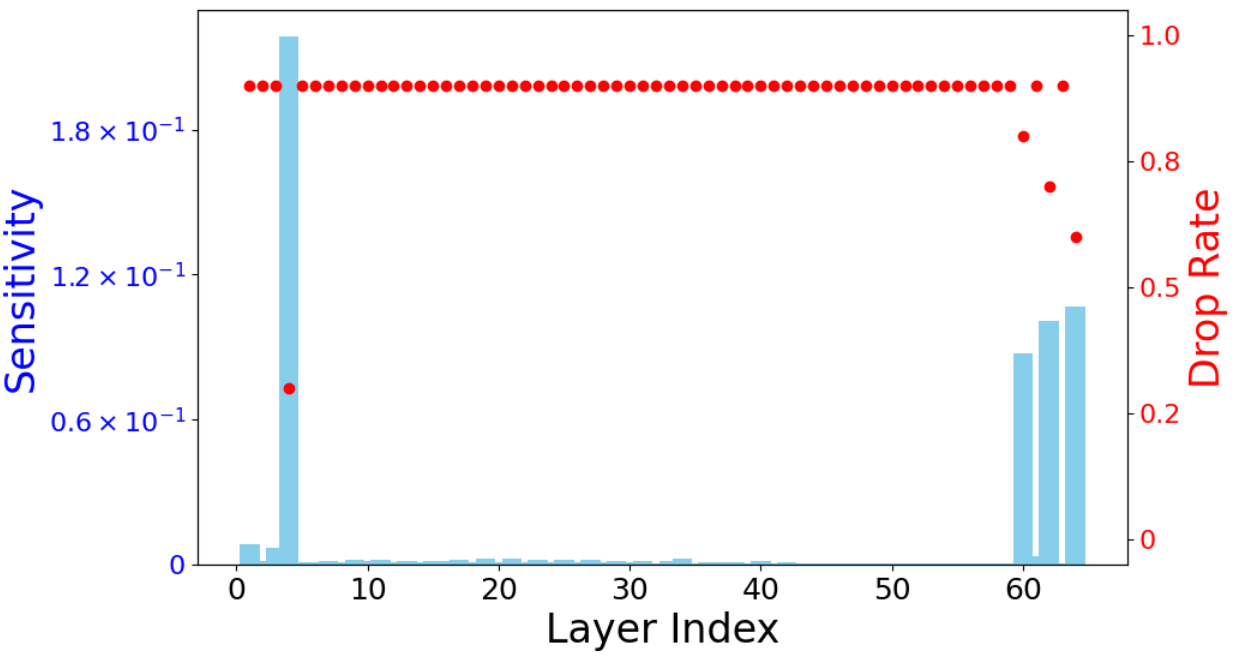}}
\end{figure}

\begin{figure}[H]
\ContinuedFloat
\subfloat[LLaMA2-70B w/ QLoRA + DropBP (p=0.5)]{\includegraphics[width=.45\textwidth]{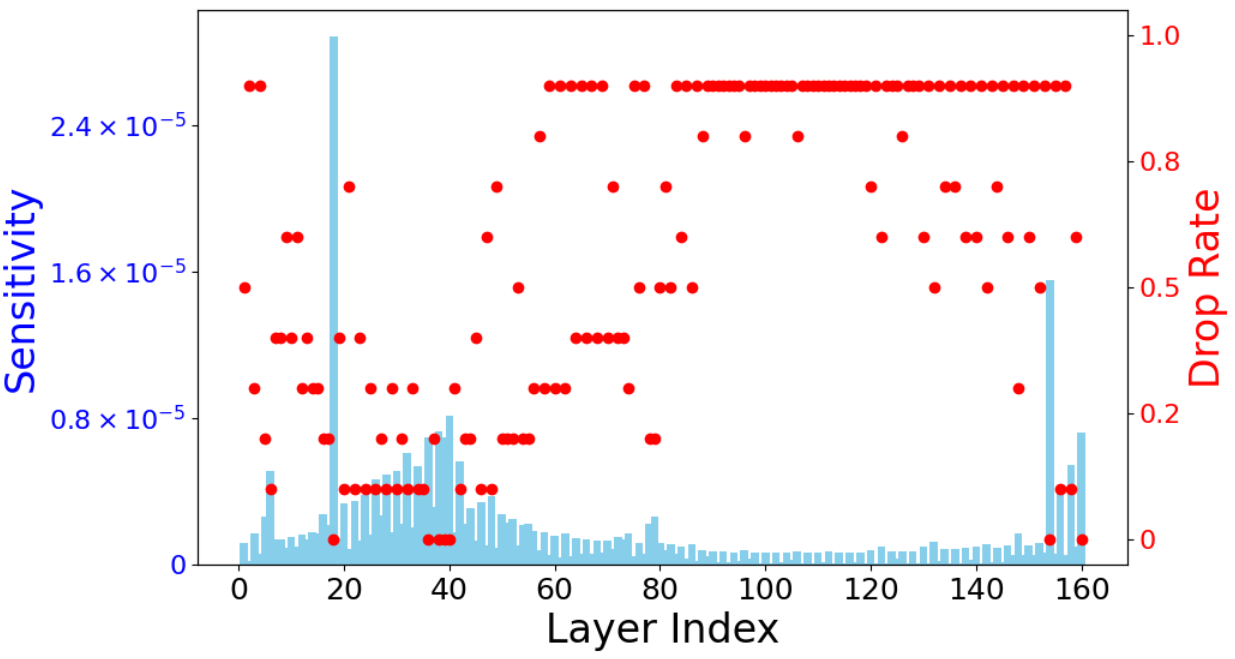}}\qquad 
\subfloat[LLaMA2-70B w/ QLoRA + DropBP (p=0.875)]{\includegraphics[width=.45\textwidth]{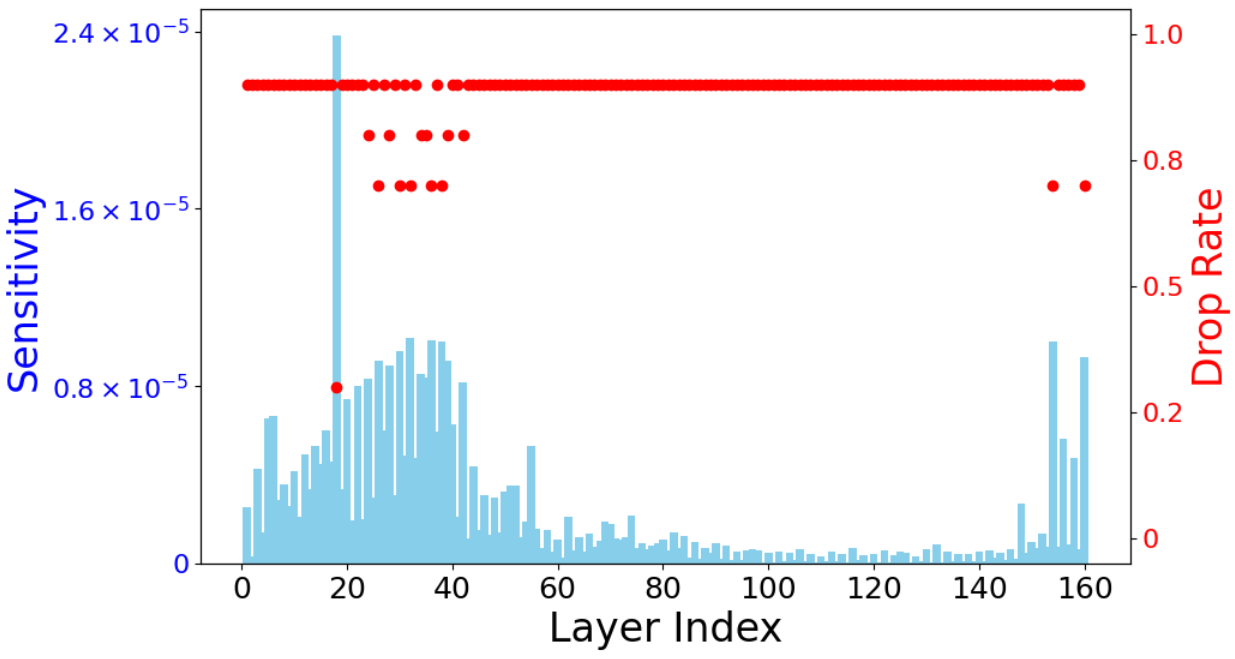}}
\caption{The distribution of drop rates determined by sensitivity when fine-tuning LLaMA2 through Full-FT, LoRA, or QLoRA using DropBP on Alpaca datasets.}
\label{fig: distribution of drop rates determined by sensitivity}
\end{figure}

\section{Comparisons between Layer Dropping and DropBP on fine-tuning LLMs}\label{appendix: comparisons between layer dropping}

In this section, we compare Layerdrop (LD) \cite{Fan-layerdrop} and Progressive Layer Dropping (PLD) \cite{Zhang-PLD} with DropBP under the same LLMs fine-tuning scenario. We set the relative FLOPs of LD and PLD to 0.75 of the baseline (LoRA), which corresponds to the same relative FLOPs when the drop rate of DropBP is set to 0.5. 

\begin{wrapfigure}{!r}{0.45\textwidth}
  \centering
  \includegraphics[width=1.0\linewidth]{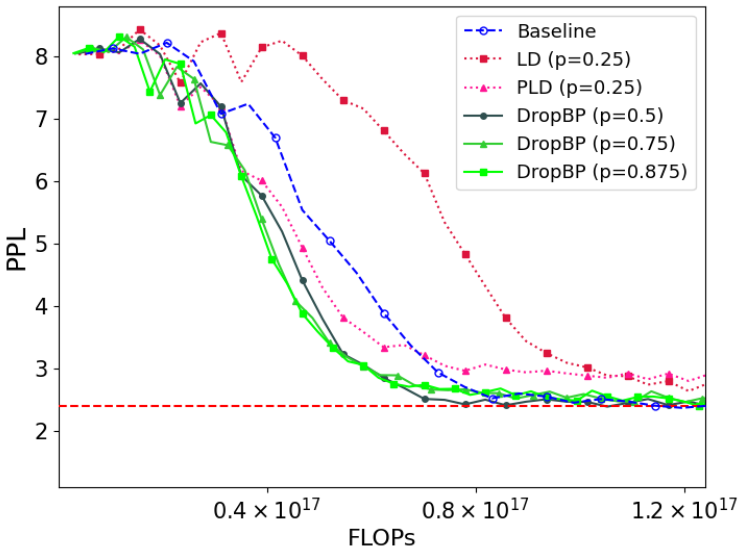}
  \caption{Validation perplexity (PPL) for fine-tuning LLaMA2-7B through LoRA (baseline) with LayerDrop (LD), Progress Layer Dropping (PLD), or DropBP on the Alpaca dataset. The $p$ represents the target average drop rate for backward propagation in DropBP.}
  \label{fig: layer dropping}
\end{wrapfigure}

As shown in Fig. \ref{fig: layer dropping}, our DropBP converges faster to the same validation PPL compared to LD and PLD. Moreover, as seen in Table \ref{tb: layer dropping results}, DropBP achieves comparable accuracy to the baseline even with a relative FLOPs of 0.56, whereas LD and PLD experience a significant accuracy drop of over 5\% with a relative FLOPs of 0.75. We believe this difference arises from the high sensitivity of forward propagation throughout the fine-tuning process. Specifically, layer dropping techniques omit certain layers of well-pretrained LLMs during forward propagation, resulting in significant output deviations that adversely impact the loss and the overall training process. Conversely, DropBP maintains all layers during forward propagation, thereby ensuring precise outputs and loss calculations, which facilitate stable training. Please note that, as explained in Section \ref{Section: related works}, LD and PLD are designed to accelerate the pretraining of small transformer models (SLMs) like BERT by dropping layers throughout the entire training process while DropBP only focuses on fine-tuning LLMs. In future studies, we will explore whether DropBP can similarly accelerate the pretraining of transformer models and investigate ways to improve its effectiveness.

\begin{table}[!ht]
\caption{Test accuracy on the 0-shot commonsense reasoning tasks when fine-tuning LLaMA2-7B through LoRA with layerdrop (LD), progressive layer dropping (PLD), and DropBP.}
\label{tb: layer dropping results}
\centering
\small
\renewcommand{\arraystretch}{1.2}
\begin{tabular}{c|c:cc:ccc}
\Xhline{3\arrayrulewidth}
\multirow{2}{*}{\textbf{Method}} & \multirow{2}{*}{\textbf{LoRA (baseline)}} & \multirow{2}{*}{\textbf{LoRA+LD}} & \multirow{2}{*}{\textbf{LoRA+PLD}} & \multicolumn{3}{c}{\textbf{LoRA+DropBP}}            \\ \cdashline{5-7}[1pt/1pt] 
                                 &                                &                                   &                                    & \textbf{p=0.5} & \textbf{p=0.75} & \textbf{p=0.875} \\ \hline
\textbf{Relative FLOPs}          & 1.00                           & 0.75                              & 0.75                               & 0.75           & 0.63            & \textbf{0.56}    \\
\textbf{Accuracy (\%)}               & 66.0                           & 58.7                              & 61.0                               & 65.9           & 66.1            & \textbf{66.4}    \\ \Xhline{3\arrayrulewidth}
\end{tabular}
\end{table}

\section{Experimental Details}\label{appendix: experimental details}

In our experimental setup, the AdamW \cite{Loshchilov-AdamW} optimizer and a cosine annealing learning rate scheduler \cite{Loshchilov-cosineannealing} were utilized as common settings. LoRA \cite{Hu-LoRA} and QLoRA \cite{Dettmers-QLoRA} were integrated to every linear layer of our model, with the LoRA parameters $r$ and $\alpha$ set to 8 and 16, respectively. We experimented with all the learning rates presented in Table \ref{tb: set up} and reported the best accuracy achieved in Table \ref{tb: main results}-\ref{tb: mt-bench}. 

\begin{table}[!ht]
\caption{Detailed Setup for Table \ref{tb: main results}-\ref{tb: mt-bench}. BS and MBS are denoted as the batch size and micro batch size, respectively. Mixed refers to mixed precision training \cite{micikevicius-AMP} using BFloat16 (BF16) and 32-bit.}
\label{tb: set up}
\centering
\small
\renewcommand{\arraystretch}{1.1}
\begin{tabular}{c|ccccccc}
\Xhline{3\arrayrulewidth}
                            & Fine-tuning & Dataset & \# Iterations & BS  & MBS & Precision & Learning rate \\ \Xhline{3\arrayrulewidth}
\multirow{4}{*}{LLaMA2-7B}  & \multirow{2}{*}{LoRA}        & Alpaca  & 25K           & 128 & 2   & Mixed     & 1e-4, 3e-4    \\
                            &             & Dolly   & 7K            & 128 & 2   & Mixed     & 1e-4, 3e-4    \\ \cdashline{2-8}[1pt/1pt]
                            & \multirow{2}{*}{Full-FT}         & Alpaca  & 25K           & 128 & 2   & BF16      & 1e-4, 3e-4    \\
                            &             & Dolly   & 7K            & 128 & 2   & BF16      & 1e-4, 3e-4    \\ \Xhline{1\arrayrulewidth}
\multirow{2}{*}{LLaMA2-13B} & \multirow{2}{*}{LoRA}        & Alpaca  & 25K           & 128 & 2   & BF16      & 1e-4, 3e-4    \\
                            &             & Dolly   & 7K            & 128 & 2   & BF16      & 1e-4, 3e-4    \\ \Xhline{1\arrayrulewidth}
\multirow{2}{*}{LLaMA-30B}  & \multirow{2}{*}{QLoRA}       & Alpaca  & 25K           & 128 & 2   & BF16      & 1e-4, 3e-4    \\
                            &             & Dolly   & 7K            & 128 & 2   & BF16      & 1e-4, 3e-4    \\ \Xhline{1\arrayrulewidth}
\multirow{2}{*}{LLaMA2-70B}  & \multirow{2}{*}{QLoRA}       & Alpaca  & 50K           & 128 & 1   & BF16      & 5e-5, 1e-4    \\
                            &             & Dolly   & 14K           & 128 & 1   & BF16      & 5e-5, 1e-4\\ \Xhline{1\arrayrulewidth}
LLaMA3-8B  & LoRA       & Oasst1  & 2.5K           & 16 & 4   & BF16      & 3e-4, 5e-4    \\ \Xhline{3\arrayrulewidth}
\end{tabular}
\end{table}


\newpage
\section*{NeurIPS Paper Checklist}

\begin{enumerate}

\item {\bf Claims}
    \item[] Question: Do the main claims made in the abstract and introduction accurately reflect the paper's contributions and scope?
    \item[] Answer: \answerYes{} 
    \item[] Justification: We appropriately present the contributions of the paper in the Abstract and Section \ref{Section: Introduction}.

\item {\bf Limitations}
    \item[] Question: Does the paper discuss the limitations of the work performed by the authors?
    \item[] Answer: \answerYes{} 
    \item[] Justification: In Section \ref{Section: related works} and Appendix \ref{appendix: comparisons between layer dropping}, we clearly state that DropBP is developed specifically for fine-tuning, unlike other layer dropping techniques, and we indicate plans for further improvements in future research.

\item {\bf Theory Assumptions and Proofs}
    \item[] Question: For each theoretical result, does the paper provide the full set of assumptions and a complete (and correct) proof?
    \item[] Answer: \answerYes{} 
    \item[] Justification: In Appendix \ref{appendix : training time reduction using dropbp}, we mathematically calculate the theoretical reduction in FLOPs achieved by DropBP. Through experiments presented in Section \ref{Subsection: main results} and Appendix \ref{appendix : training time reduction using dropbp}, we confirm that this theoretical reduction closely matches the actual training time reduction.

    \item {\bf Experimental Result Reproducibility}
    \item[] Question: Does the paper fully disclose all the information needed to reproduce the main experimental results of the paper to the extent that it affects the main claims and/or conclusions of the paper (regardless of whether the code and data are provided or not)?
    \item[] Answer: \answerYes{} 
    \item[] Justification: In Section \ref{Subsection: Implementation} and Appendix \ref{appendix: experimental details}, we provide detailed information to ensure reproducibility, and in the abstract, we present the anonymous code in Abstract to implement this.

\item {\bf Open access to data and code}
    \item[] Question: Does the paper provide open access to the data and code, with sufficient instructions to faithfully reproduce the main experimental results, as described in supplemental material?
    \item[] Answer: \answerYes{} 
    \item[] Justification: We present the anonymous code for reproducibility in the Abstract.

\item {\bf Experimental Setting/Details}
    \item[] Question: Does the paper specify all the training and test details (e.g., data splits, hyperparameters, how they were chosen, type of optimizer, etc.) necessary to understand the results?
    \item[] Answer: \answerYes{} 
    \item[] Justification: We provide the experimental details in Section \ref{Subsection: Implementation}, Appendix \ref{appendix: experimental details}, and the anonymous code presented in the Abstract.

\item {\bf Experiment Statistical Significance}
    \item[] Question: Does the paper report error bars suitably and correctly defined or other appropriate information about the statistical significance of the experiments?
    \item[] Answer: \answerNo{} 
    \item[] Justification: Although we conducted experiments with multiple seeds and report the average values, we did not include error bars or statistical information because the deviations were minimal, and including them would detract from the clarity of the paper.

\item {\bf Experiments Compute Resources}
    \item[] Question: For each experiment, does the paper provide sufficient information on the computer resources (type of compute workers, memory, time of execution) needed to reproduce the experiments?
    \item[] Answer: \answerYes{} 
    \item[] Justification: We mention the types of devices used for the experiments in Section \ref{Subsection: Implementation}, and provide the time required to reproduce the experimental results in Section \ref{Subsection: main results}.
    
\item {\bf Code Of Ethics}
    \item[] Question: Does the research conducted in the paper conform, in every respect, with the NeurIPS Code of Ethics \url{https://neurips.cc/public/EthicsGuidelines}?
    \item[] Answer: \answerYes{}
    \item[] Justification: Our paper adheres to the NeurIPS Code of Ethics.

\item {\bf Broader Impacts}
    \item[] Question: Does the paper discuss both potential positive societal impacts and negative societal impacts of the work performed?
    \item[] Answer: \answerNA{} 
    \item[] Justification: Since we discuss a efficient fine-tuning algorithm for LLMs, we believe that our work does not have any direct negative societal impact.
    
\item {\bf Safeguards}
    \item[] Question: Does the paper describe safeguards that have been put in place for responsible release of data or models that have a high risk for misuse (e.g., pretrained language models, image generators, or scraped datasets)?
    \item[] Answer: \answerNA{} 
    \item[] Justification: Our work is related to the efficient fine-tuning of LLMs, and therefore, there are no specific safeguards described for the responsible release of data or models, as the nature of our research does not involve high-risk misuse scenarios.

\item {\bf Licenses for existing assets}
    \item[] Question: Are the creators or original owners of assets (e.g., code, data, models), used in the paper, properly credited and are the license and terms of use explicitly mentioned and properly respected?
    \item[] Answer: \answerYes{} 
    \item[] Justification: We appropriately cite the original paper and existing codes in Section \ref{Subsection: Implementation} and our code in Abstract.

\item {\bf New Assets}
    \item[] Question: Are new assets introduced in the paper well documented and is the documentation provided alongside the assets?
    \item[] Answer: \answerYes{} 
    \item[] Justification: We provide the new code with a proper license as an anonymized URL in the Appendix.

\item {\bf Crowdsourcing and Research with Human Subjects}
    \item[] Question: For crowdsourcing experiments and research with human subjects, does the paper include the full text of instructions given to participants and screenshots, if applicable, as well as details about compensation (if any)? 
    \item[] Answer: \answerNA{} 
    \item[] Justification: Our work does not involve crowdsourcing nor research with human subjects.

\item {\bf Institutional Review Board (IRB) Approvals or Equivalent for Research with Human Subjects}
    \item[] Question: Does the paper describe potential risks incurred by study participants, whether such risks were disclosed to the subjects, and whether Institutional Review Board (IRB) approvals (or an equivalent approval/review based on the requirements of your country or institution) were obtained?
    \item[] Answer: \answerNA{} 
    \item[] Justification: Our work does not involve crowdsourcing nor research with human subjects.

\end{enumerate}
\end{document}